\documentclass[10pt,journal,compsoc]{IEEEtran}
\usepackage{comment}
\usepackage{amsmath,amssymb} 
\usepackage{bbding}
\usepackage{array}
\usepackage{mathtools} 
\usepackage{bbding}
\usepackage{multirow}
\usepackage{utfsym}
\usepackage{booktabs}
\usepackage{colortbl}

\usepackage[breaklinks=true,colorlinks,bookmarks=false]{hyperref}
\newlength\savewidth

\ifCLASSOPTIONcompsoc
  \usepackage[nocompress]{cite}
\else
  \usepackage{cite}
\fi

\usepackage{amsmath}
\usepackage{algorithm}
\usepackage{algorithmic}
\usepackage{array}
\ifCLASSOPTIONcompsoc
 \usepackage[caption=false,font=footnotesize,labelfont=sf,textfont=sf,position=top]{subfig}
\else
 \usepackage[caption=false,font=footnotesize]{subfig}
\fi

\usepackage{url}
\begin{document}
%
\title{iSeg: An Iterative Refinement-based Framework for Training-free Segmentation}
%
%
%
%

\author{Lin Sun,
        Jiale Cao,
        Jin Xie,
        Fahad Shahbaz Khan,~\IEEEmembership{Senior Member,~IEEE,}\\
        and~Yanwei~Pang,~\IEEEmembership{Senior Member,~IEEE}
\IEEEcompsocitemizethanks{
\IEEEcompsocthanksitem L. Sun, J. Cao, and Y. Pang are with the School of Electrical and Information Engineering, Tianjin University, Tianjin 300072, China, and also with the Shanghai Artificial Intelligence Laboratory, Shanghai 200232, China. (E-mail: \{sun0806,~connor,~pyw\}@tju.edu.cn) 
\IEEEcompsocthanksitem J. Xie is is with School of Big Data and Software Engineering, Chongqing University, Chongqing 401331, China. (E-mail: xiejin@cqu.edu.cn)
\IEEEcompsocthanksitem F. Khan are both with  Mohamed bin Zayed University of Artificial Intelligence, UAE, and also with  Linkoping University, Sweden. (E-mail: fahad.khan@mbzuai.ac.ae)
}
}
%
%

\markboth{Submitted to IEEE Transactions on Pattern Analysis and Machine Intelligence}%
{Shell \MakeLowercase{\textit{et al.}}: Bare Demo of IEEEtran.cls for Computer Society Journals}
\IEEEtitleabstractindextext{%
\begin{abstract}
Stable diffusion has demonstrated strong image synthesis ability to given text descriptions, suggesting it to contain strong semantic clue for grouping  objects. The researchers have explored employing stable diffusion for training-free segmentation. 
Most existing approaches  refine cross-attention map by self-attention map once, demonstrating that self-attention map contains useful semantic information to improve segmentation. To fully utilize self-attention map, we present a deep experimental analysis on iteratively refining cross-attention map with self-attention map, and propose an effective iterative refinement framework for training-free segmentation, named iSeg.
The proposed iSeg introduces an entropy-reduced self-attention module that utilizes a gradient descent scheme to reduce the entropy of self-attention map, thereby suppressing the weak responses corresponding to irrelevant global information. Leveraging the entropy-reduced self-attention module, our iSeg stably improves refined cross-attention map with iterative refinement. Further, we design a category-enhanced cross-attention module to generate accurate cross-attention map, providing a better initial input for iterative refinement. Extensive experiments across different datasets and diverse segmentation tasks (weakly-supervised semantic segmentation, open-vocabulary semantic segmentation, unsupervised segmentation, and mask generation on synthetic dataset) reveal the merits of proposed contributions, leading to promising performance on  diverse segmentation tasks. For unsupervised semantic segmentation on Cityscapes, our iSeg achieves an absolute gain of $3.8\%$ in terms of mIoU compared to the best existing training-free approach in literature. Moreover, our proposed iSeg can support segmentation with different kinds of images and interactions. 
The project  is available at \url{https://linsun449.github.io/iSeg}.

\end{abstract}


}

\maketitle

\IEEEdisplaynontitleabstractindextext

%
\IEEEpeerreviewmaketitle

\IEEEraisesectionheading{\section{Introduction}\label{secIntroduction}}


Semantic segmentation \cite{Chen_2017_DeepLabV3, Wang_2021_MaxDeepLab, Xie_2021_SegFormer} aims to differentiate the pixels into different semantic groups.

In past decade, fully-supervised semantic segmentation has achieved significant progress where the models are trained on large-scale dataset with pixel-level annotations and fixed  categories.  This fully-supervised setting hinders zero-shot segmentation ability on unseen categories. Further, it is expensive to collect and annotate large-scale dataset with numerous categories \cite{Lin_2019_Block} for model training. One straightforward solution is to develop a universal and training-free framework that exhibits strong generalization capabilities across arbitrary categories without any training on segmentation datasets.

Recent work of \cite{Radford_2021_CLIP} introduces a vision-language model, CLIP, that learns the connection between text and image through training on large-scale image-text pairs, showing promising generalization capabilities in recognizing arbitrary categories. Inspired by this, several works \cite{Liang_2023_OVSeg, Lin_2023_CLIPES, Xu_2022GroupVit, Xie_2022_Clims} explore employing the strong zero-shot ability of CLIP for semantic segmentation. Some works focus on directly utilizing the learned text-image mapping from pre-trained CLIP. For instance, OVSeg \cite{Liang_2023_OVSeg} first utilizes mask proposal generator to extract some class-agnostic masks, and then employs CLIP to classify the cropped and masked images for open-vocabulary semantic segmentation. 
Some works employ text supervision similar to CLIP. For instance, GroupViT \cite{Xu_2022GroupVit} introduces a hierarchical grouping vision transformer with text supervision, which is able to group different semantic regions without pixel-level annotations. 
These methods are able to improve the generalization ability of semantic segmentation or reduce the reliance on pixel-level annotations in some degree. However,  CLIP focuses on image-level information, limiting its capability in pixel-level segmentation tasks, especially under training-free setting.

\begin{figure*}[t]
\centering
\footnotesize
\includegraphics[width=0.98\linewidth]{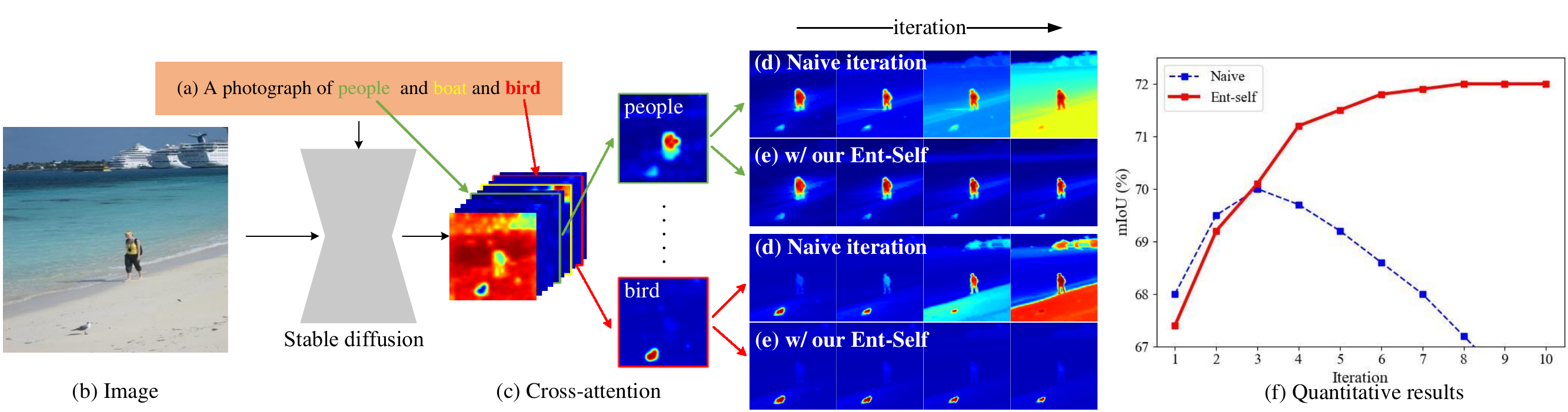}
\caption{\textbf{Comparison of naive iteration strategy and our entropy-reduced self-attention (Ent-Self) module.} We feed text prompt (a) and image (b) into pre-trained stable diffusion  to extract cross-attention (c) and self-attention maps. We only select  cross-attention maps corresponding to category names, and do not show self-attention map here for simplicity. Then, we refine these selected cross-attention maps  with self-attention map using naive iteration strategy and our Ent-Self.  Naive iteration strategy directly refines cross-attention maps using original self-attention map, which leads to noisy cross-attention maps  after multiple iterations, such as the feature maps of  categories `people' and `bird' in (d). Compared to naive iteration, our Ent-Self generates accurate refined cross-attention maps of categories `people' and `bird' in (e). In (f), we give  quantitative comparison on  pseudo mask generation of PASCAL VOC in weakly-supervised semantic segmentation. Compared to naive iteration, using our Ent-Self can improve mask generation with the increment of iterations.}
\label{fig:09introduction}
\end{figure*}

Recently, researchers started to employ text-to-image stable diffusion model for segmentation. Stable diffusion \cite{Rombach_2021_LDM} was initially designed to perform high-quality image synthesis. With the pre-training on large-scale text-image paired data, stable diffusion is able to generate high-quality images with reasonable structure according to given text descriptions. Inspired by this, researchers believed that stable diffusion contains rich semantic clue to group objects, and proposed various diffusion-based segmentation approaches. Some approaches \cite{Wolleb_2022_DIISE, Baranchuk_2021_labelefficient, Zhao_2023_VPD, Wan_2023_PMP} treat stable diffusion as feature extractor, and re-train it on segmentation dataset. In contrast, other approaches \cite{Tian_2023_DiffSeg, Wang_2023_DiffSegmenter, Xiao_2023_T2M} explore training-free framework by utilizing self-attention map and cross-attention map in stable diffusion. 
For instance, DiffSeg \cite{Tian_2023_DiffSeg} iteratively merges self-attention maps at different positions and employs non-maximum suppression to assign the pixels into different objects. DiffuMask \cite{Wu_2023DiffuMask} generates initial object masks of synthetic images using cross-attention maps. DiffSegmenter \cite{Wang_2023_DiffSegmenter} refines the cross-attention map with self-attention map for improved segmentation, while Dataset Diffusion \cite{Nguyen_2023_DiffusionDataset} refines the cross-attention map using self-attention exponentiation.  These approaches present an initial attempt on employing self-attention and cross-attention maps for training-free segmentation tasks, demonstrating that self-attention and cross-attention maps contain useful semantic information to improve segmentation. Therefore, there is a natural question that, how to fully harness self-attention and cross-attention maps for segmentation tasks.

In this paper, to fully exploit the potential ability of cross-attention and self-attention maps in pre-trained stable diffusion  for training-free segmentation, we conduct a preliminary experiment, where we iteratively refine the cross-attention map with  self-attention map as in Fig. \ref{fig:09introduction}. Given a text prompt (a) and an image (b), we employ pre-trained stable diffusion  to extract cross-attention maps and select the cross-attention maps corresponding to different category names, such as people and bird, in (c). Subsequently, we perform a naive iteration strategy to refine the cross-attention maps by iteratively multiplying self-attention map. After several iterations, we observe that the refined cross-attention maps, associated with categories people and bird, gradually have strong response on other regions in (d). We argue that this degradation in refined cross-attention maps is  caused by self-attention map that aggregates global information from irrelevant regions after multiple iterations. 

To address this issue mentioned above, we propose an iterative refinement framework  by introducing an entropy-reduced self-attention (Ent-Self) module to impede the information aggregation from irrelevant regions. As in (e), using our Ent-Self module is able to generate the stable and accurate cross-attention maps of person and bird after multiple iterations. We further give a quantitative comparison of generated pseudo mask for weakly-supervised semantic segmentation in (f). Compared to naive iteration strategy, using our Ent-Self module exhibits improved performance with the increment of iterations. In addition, we design a simple category-enhanced cross-attention (Cat-Cross) module to generate more accurate cross-attention maps corresponding to the category names, which are  fed to iterative refinement. The Cat-Cross module employs a weighting strategy to enhance the text embeddings of given category names while suppressing the text embeddings of other words. We perform the experiments on various datasets and  segmentation tasks, including weakly-supervised semantic segmentation, open-vocabulary semantic segmentation, unsupervised segmentation, and mask generation on synthetic dataset. 
 Extensive experiments 
 demonstrate the efficacy of our proposed method. 
The contributions and merits of our method can be summarized as
\begin{itemize}
    \item We give a deep analysis on simply refining cross-attention map using self-attention map with multiple iterations, and Analyze  the reason why this naive iteration strategy presents worse performance. 
    
    \item  We introduce an effective iterative refinement for improved segmentation using our proposed entropy-reduced self-attention module, which can avoid incorporate irrelevant global information within self-attention maps during multiple iterations. 
    
    \item  A category-enhanced cross-attention map is proposed to enhance the text embeddings of given categories while suppressing other words.

    \item  The experiments are performed on various segmentation tasks, including weakly-supervised semantic segmentation, open-vocabulary semantic segmentation, unsupervised segmentation, and mask generation on synthetic dataset. Our proposed method can significantly improve segmentation accuracy.
    
    \item  Our proposed method can  support the segmentation on different kinds of images or different interactions, such as point, line and box.    
    
\end{itemize}

\section{Related Work}
In this section, we first give a review on semantic segmentation. Afterwards, we introduce the related works of using diffusion models for segmentation.

\subsection{Semantic Segmentation}
Semantic segmentation is one of classical computer vision tasks, which aims to group the pixels belonging to same semantic category together. In the past decade, semantic segmentation has achieved the great progress. The related methods  can be mainly divided into CNN-based approaches and transformer-based approaches.  

Initially, the researchers mainly investigated CNN-based networks. One of the pioneering approaches is FCN \cite{Long_2015_FCN}, which utilizes the convolutional neural networks with skip-layer connections for improved semantic segmentation. Subsequently, numerous variants were developed, incorporating innovations such as spatial pyramid structures \cite{Chen_2017_DeepLabV3, Chen_2018_DeepLabV2}, multi-layer feature fusion \cite{Liu_2016_SSD, Badrinarayanan_2017_SegNet, Chen_2018_Encoder}, and non-local enhancements \cite{Fu_2019_DANet, Huang_2019_CCNet, Yuan_2021_OCNet}. The spatial pyramid structure usually adopts multiple paralleled branches with different receptive fields to extract diverse context information. Multi-layer feature fusion employs skip-layer connection to fuse the feature maps of different layers. Non-local enhancement adopts the attention operation to integrate globally context information.
With the success of vision transformer in image classification \cite{Dosovitskiy_2021_VIT, Carion_2020_End}, researchers began to explore transformer-based segmentation approaches. Some studies \cite{Yuan_2021_HRFormer, Gu_2022_HRViT, Xie_2021_SegFormer} focus on harnessing the powerful representational capabilities of transformer by employing it as backbone. Additionally, other researches \cite{Cheng_2021_Perpixel, Cheng_2022_Masked} conceptualize semantic segmentation as a set prediction task, utilizing the transformer as the decoder.

These approaches mentioned above belong to fully-supervised semantic segmentation. 
Besides, researchers developed various semantic segmentation paradigms, such as weakly-supervised semantic segmentation, open-vocabulary semantic segmentation, and unsupervised segmentation. Compared to fully-supervised semantic segmentation, these tasks are more challenging. Weakly-supervised semantic segmentation \cite{Papandreou_2015_Weakly, Pinheiro_2015_I2P, Ahn_2018_Affinity} solely relies on image-level category annotations to train the models. For this task, many researchers employed CAM \cite{Zhou_2016_CAM} to generate pseudo pixel-level  mask annotations and then trained the models with these pseudo pixel-level  annotations. Open-vocabulary semantic segmentation \cite{Li_2022_Language, Ghiasi_2022_Scaling, Xu_2022GroupVit} aims to perform semantic segmentation on unseen categories. In open-vocabulary semantic segmentation, researchers mainly explored to utilize the pre-trained CLIP \cite{Radford_2021_CLIP} to segment unseen categories. Unsupervised segmentation \cite{Ji_2019_IIC, Hamilton_2022_STEGO, Shin_2022_RECO} aims to group the pixels belonging to the same objects without using any annotation information.

\subsection{Diffusion Models for Segmentation}
Diffusion models \cite{Dhariwal_2021_DPM} have exhibited strong image synthesis ability across various text prompts, indicating their strong semantic clue for image segmentation. Inspired by this, researchers started to explore employing the pre-trained diffusion models for segmentation task. For instance, VPD \cite{Zhao_2023_VPD} views the pre-trained stable diffusion model as feature extractor, and fine-tunes it for semantic segmentation and depth estimation tasks. TADP \cite{Kondapaneni_2023_TADP} introduces image-aligned text prompts to condition the feature extraction of stable diffusion model for semantic segmentation. SLIME \cite{Khani_2023_Slime} adopts stable diffusion model with learnable text embeddings for few-shot segmentation. Additionally, some researchers \cite{Ji_2023_DDP}  adopted diffusion framework, in which the model aims to  predict the accurate segmentation maps from noisy data using denoising process.
  
Recently, some researchers explored training-free diffusion models for segmentation.  One of the key idea is to employ self-attention map and cross-attention map from pre-trained diffusion model to generate masks of different  categories. For instance, DiffuMask \cite{Wu_2023DiffuMask} first employs diffusion model to generate some synthetic images, and calculates the averaged cross-attention maps as the initial object masks which are refined by AffinityNet. DiffSeg \cite{Tian_2023_DiffSeg} explores grouping self-attention maps at different positions into distinct objects based on the similarities of self-attention maps, followed by a non-maximum suppression operation to assign the pixels to individual objects. OVDiff \cite{Karazija_2023_OVDiff} generates synthetic images by diffusion model, and uses a feature extractor to generate the foreground and background prototypes of different categories for open-vocabulary segmentation. 
Both DiffSegmenter \cite{Wang_2023_DiffSegmenter} and T2M \cite{Xiao_2023_T2M} refine cross-attention maps of different categories with self-attention maps. Dataset Diffusion  \cite{Nguyen_2023_DiffusionDataset} introduces a novel self-attention exponentiation that performs power operation on self-attention maps before refining cross-attention maps. In some degree, self-attention exponentiation   can be viewed as a naive refinement as in Section Introduction. Compared to our method, naive refinement can only present improvement at few iterations as in Fig. \ref{fig:09introduction}(f). To improve cross-attention map, DiffSegmenter, Dataset Diffusion, and T2M employ additional BLIP \cite{Li_2022_BLIP}, ChatGPT, and CLIP to carefully design text prompts, which are relatively complex in some degree. Though these aforementioned approaches have exploited self-attention and cross-attention maps for training-free segmentation, we argue that these approaches do not fully utilize the semantic clue existing in self-attention and cross-attention maps.
In this paper, we first provide a deep analysis on refining cross-attention map with self-attention map multiple times, and then propose an effective iterative training-free segmentation framework using pre-trained stable diffusion model.


\begin{figure*}[t]
\centering
\includegraphics[width=0.98\linewidth]{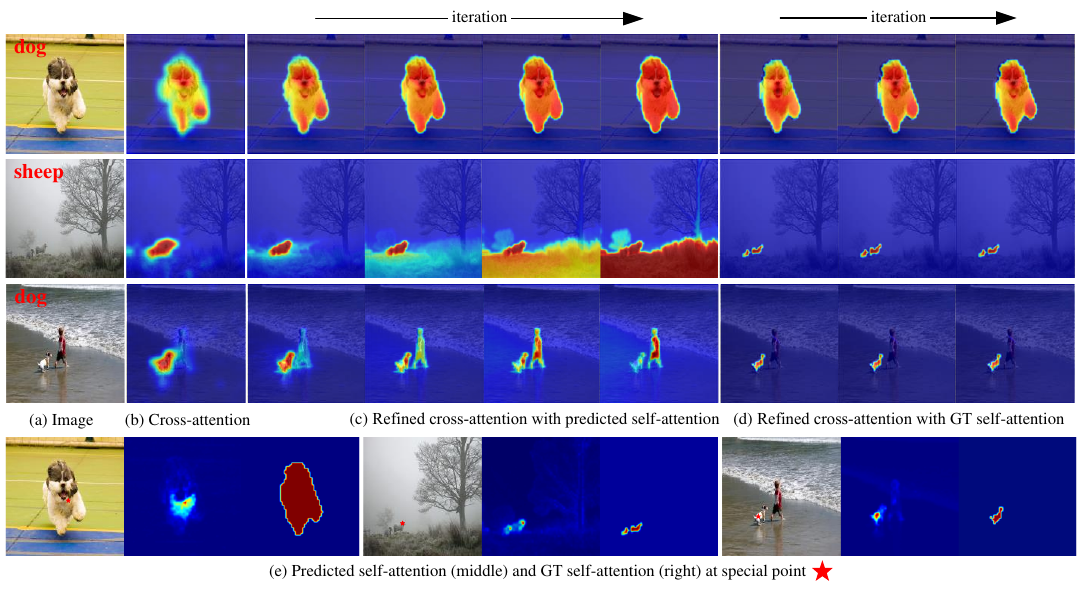}
\caption{\textbf{Cross-attention map refinement with predicted self-attention map and ground-truth self-attention map.} Given the  image (a) and cross-attention map (b), we first give the refined cross-attention maps at different iterations using predicted self-attention map in (c). At the same time, we  present the refined cross-attention maps using ground-truth (GT) self-attention map in (d), where GT self-attention map is generated by ground-truth masks. Finally, we compare the predicted self-attention map and GT self-attention map at selected point in (e), where predicted self-attention map is noisy and GT self-attention map is clean.}
\label{fig:03problem}
\end{figure*}
\section{Method}
\label{sec:ours}

In this section, we first give some preliminaries about stable diffusion model and training-free segmentation. Afterwards, we  introduce the motivation of our proposed method. Finally, we describe our proposed iterative refinement framework for training-free segmentation.

\subsection{Preliminaries}
\noindent\textbf{Stable diffusion model.} Text-to-image (T2I) diffusion model is a type of text-conditioned diffusion models \cite{Dhariwal_2021_DPM}, such as DALL-E \cite{Ramesh_2022_DALL}, GLIDE \cite{Nichol_2022_Glide}, Imagen \cite{Saharia_2022_Imagen}, and stable diffusion \cite{Rombach_2021_LDM}. As one of representative T2I models, stable diffusion can generate high-quality images with  text prompts, due to the training on large-scale text-image paired dataset (\textit{e.g.,} LAION-5B \cite{Schuhmann_2022_Laion}). It comprises three components: (1) A pre-trained VAE \cite{kingma2013vae} encoder and decoder for image compression and restoration. (2) A text encoder that encodes text descriptions into latent semantic space to condition image generation. (3) A transformer-based denoising UNet \cite{Ronneberger_2015_Unet} that recovers images from noisy data according to the given text descriptions.

\noindent\textbf{Attention modules.} Denoising UNet in stable diffusion integrates self-attention and cross-attention mechanisms. The self-attention models the relation between different spatial positions of visual features, while the cross-attention models the relation between visual features and text embeddings. The attention map in both self-attention and cross-attention operations can be calculated by the query $Q$ and the key $V$, which can be written  as 
\begin{equation}
A = \mathrm{Softmax}(\frac{QK^{\mathrm{T}}}{\sqrt{d}}),
\label{eq:attn}
\end{equation}
where $d$ is normalization factor. Assuming that the visual features have the spatial size of $H\times W$, and the text embeddings have the length of $T$. In self-attention operation, the query and key are both generated by visual features via the linear layers. The generated self-attention map is denoted as $A_\mathrm{sa} \in \mathcal{R} ^{HW \times HW}$, which can reflect spatial similarities of different positions within image. In cross-attention operation, the query and the key are respectively generated by visual features and text embeddings via the linear layers, respectively. The generated cross-attention map is denoted as $A_\mathrm{ca} \in \mathcal{R} ^{HW \times T}$, which can capture the similarities between image and text.

To achieve training-free segmentation, a straightforward way is setting the text prompts based on given category names, such as \textit{`A photograph of cat and dog'}, then performing the forward process of stable diffusion, and finally generating the corresponding cross-attention map $A_\mathrm{ca}$ between visual features and text embeddings. Afterwards, we can predict segmentation map  by selecting the cross-attention map corresponding to the category name. Instead of directly using the  cross-attention map, some researchers \cite{Khani_2023_Slime,Nguyen_2023_DiffusionDataset} proposed to generate refined cross-attention map $A_{\mathrm{ca}}^{1}$ with self-attention map, which can be written as  
\begin{equation}
A_{\mathrm{ca}}^1  = A_{\mathrm{sa}} \times A_{\mathrm{ca}}.
\label{eq:refined}
\end{equation}
Compared to original cross-attention map, the refined cross-attention map incorporates spatial similarities between different positions existing in self-attention map \cite{Khani_2023_Slime}, which is useful to perform accurate segmentation. 

\begin{figure*}[t]
\centering
\includegraphics[width=0.98\linewidth]{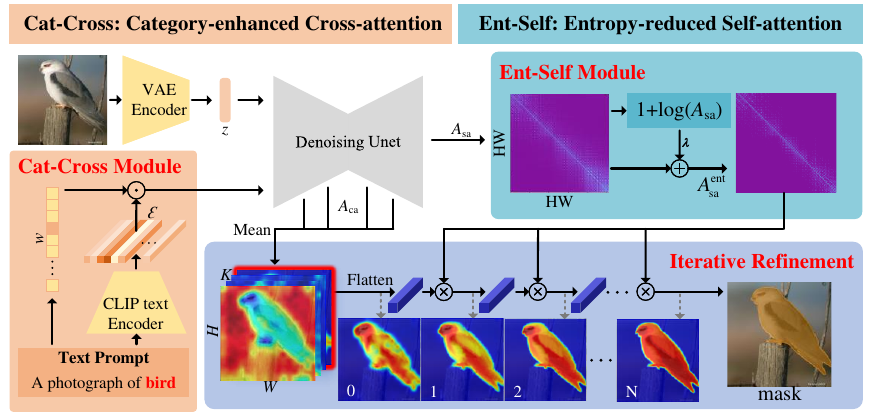}
\caption{\textbf{Architecture of our proposed training-free iSeg.} iSeg comprises two novel modules. The Cat-Cross module generates more accurate cross-attention map, while the Ent-Self module reduces irrelevant information in self-attention map. Given an image with paired text prompt, we first get  latent feature $z$ and embedding feature $\varepsilon$ by VAE encoder and Cat-Cross module respectively. These features are fed into  denoising U-net to extract cross-attention map $A_{\mathrm{ca}}$ and self-attention map $A_{\mathrm{sa}}$. Then the Ent-Self module is applied to reduce entropy of $A_{\mathrm{sa}}$ and obtain $A_{\mathrm{sa}}^{\mathrm{ent}}$. Finally, an iterative refinement is conducted to refine cross-attention map with entropy-reduced self-attention map.}
\label{fig:00arch}
\end{figure*}

\subsection{Motivation}
\noindent\textbf{Problem.} As mentioned above, self-attention map can be used to refine the cross-attention map. Therefore, a natural question is that, can we improve cross-attention map iteratively with self-attention map, formulated as $A_{\mathrm{ca}}^{n}  = A_{\mathrm{sa}} \times A_{\mathrm{ca}}^{n-1}$, where $n$ is the index of iterations. We normalize the refined cross-attention map after each iteration. We call this strategy as naive iterative refinement. 
We initially believe that this naive iterative refinement would yield improved segmentation performance. However, as in Fig. \ref{fig:09introduction}(f), it presents performance degradation after several iterations. To better analyze this question, we perform the experiment in Fig. \ref{fig:03problem}. Given the input image (a) and cross-attention map (b), we present the refined cross-attention map at different iterations in (c). It can be observed that the refinement through multiple iterations can not guarantee to generate better refined cross-attention map. 
 We divide the refined results into three situations. The first situation is that the cross-attention map becomes better with the increasing iterations, such as cross-attention map of dog in first row. The second one is the cross-attention map becomes larger, such as the cross-attention map of sheep in the second row. The last one is that cross-attention map shifts to other regions, exemplified by the cross-attention map of dog showing strong response on the person in third row. Self-attention exponentiation \cite{Nguyen_2023_DiffusionDataset} can be also viewed as a naive  refinement in some degree, which does not provide deep analysis why it does not work with increasing iterations.

\noindent\textbf{Analysis.} Ideally, we argue the refinement with multiple iterations at least presents a stable cross-attention map around corresponding category if the self-attention map is relatively accurate. We further perform a simple experiment by refining cross-attention map with the ground-truth self-attention map in Fig. \ref{fig:03problem}(d). The ground-truth self-attention map is generated by self-correlation of ground-truth masks, which is binary and exclusively has responses within object regions. By using ground-truth self-attention map, the refined cross-attention map becomes stable and accurate after only one iteration. This observation demonstrates that an accurate self-attention map is crucial for accurately refining cross-attention map. As shown in (e), we compare the predicted self-attention map (middle) and ground-truth self-attention map (right) at a given point $\bigstar$. Compared to ground-truth self-attention map, predicted self-attention map is relatively noisy and has some weak responses at irrelevant regions. As a result, compared to one iteration, multiple iterations with predicted self-attention map aggregate much irrelevant global information into the refined cross-attention map. Therefore, we argue that it is important to reduce irrelevant global information within self-attention map for iterative refinement. To achieve this goal, we introduce an effective iterative refinement with entropy-reduced self-attention module below.

\subsection{Our iterative refinement framework}
Here we introduce our proposed iterative refinement framework for training-free segmentation, named iSeg. We employ pre-trained stable diffusion model for feature extraction, and perform segmentation using the generated self-attention and cross-attention maps with iterative refinement. To address performance degradation issue in iterative refinement, we introduce an entropy-reduced self-attention (Ent-Self) module to suppress irrelevant global information within self-attention map. We find that, our iterative refinement with entropy-reduced module can be also viewed as to generate the refined self-attention map closer to optimal solution (i.e., the ground-truth self-attention map). 
In addition, we design a category-enhanced cross-attention (Cat-Cross) module to generate more accurate cross-attention map as the input of iterative refinement.

\noindent\textbf{Iterative refinement.} Fig. \ref{fig:00arch} shows the overall architecture of our proposed iSeg. Given an image $I$, we feed it into a VAE encoder to extract visual latent features $z$. At the same time, we employ a text encoder to generate text embeddings $\varepsilon$ from corresponding input texts. Afterwards, we feed the visual latent features and text embeddings into the denoising UNet, which is formulated as 
\begin{equation}
n_{\mathrm{p}}=\mu(z,\varepsilon,t),
\label{eq:unet}
\end{equation}
where $\mu$ represents the denoising UNet, and $t$ indicates the timestep that is set as a fixed value $100$ like \cite{Wang_2023_DiffSegmenter}. Then, we can extract self-attention and cross-attention maps at different layers of denoising UNet. Similar to \cite{Khani_2023_Slime}, we only select the high-resolution self-attention map at last decoder layer, and upsample and fuse the cross-attention maps at different decoder layers as initial cross-attention map $A_{\mathrm{ca}}$.
For self-attention map $A_{\mathrm{sa}}$, we introduce an entropy-reduced self-attention (Ent-Self) module to reduce irrelevant global information within original self-attention map, where the entropy-reduced self-attention map is represent as $A_{\mathrm{sa}}^{\mathrm{ent}}$.  Then, we perform iterative refinement using entropy-reduced self-attention map as 
\begin{equation}
A_{\mathrm{ca}}^{n} = A_{\mathrm{sa}}^{\mathrm{ent}} \ast A_{\mathrm{ca}}^{n-1}, n=1,...,N,
\label{eq:cws}
\end{equation}
where $N$ represents the total number of iterations. $A_{\mathrm{ca}}^{0}$ is the original cross-attention map corresponding to category names, which is selected from initial cross-attention map $A_{\mathrm{ca}}$ according to the category indices. After each iteration, we add a normalization operation. For simplicity, we do not show the normalization in Eq. \ref{eq:cws} and Fig. \ref{fig:00arch}. Finally, we generate the segmentation map by binarizing  the refined cross-attention map $A_{\mathrm{ca}}^{N}$.

It is worth noting that an accurate initial cross-attention map  is also important, which provides a better input for iterative refinement. Therefore, instead of original cross-attention map, we design a simple category-enhanced cross-attention (Cat-Cross) module to obtain more accurate cross-attention map fed to iterative refinement. 

\noindent\textbf{Entropy-reduced self-attention module.} As mentioned above, the weak responses of irrelevant regions within self-attention map contain the global information and interfere with the refined cross-attention map. To address this issue, we introduce an entropy-reduced self-attention (Ent-Self) module.  Following \cite{Jin_2024_ENT}, we calculate self-attention entropy by viewing the attention distribution as a probability mass function of a discrete random variable, which is written as
\begin{equation}
\mathrm{\mathit{E}} =-\sum_{i=1}^{HW}\sum_{j=1}^{HW}A_{\mathrm{sa}}[i, j] \mathrm{log}(A_{\mathrm{sa}}[i, j]).
\label{eq:entropy}
\end{equation}
The self-attention entropy $\mathrm{\mathit{E}}$ has the maximum value when $A_{\mathrm{sa}}[i, j] = \frac{1}{N}$ for all elements, and has the minimum value when $A_{\mathrm{sa}}$ is binary like ground-truth self-attention map mentioned above. This means that the higher the self-attention entropy is, the more global information will be, and thereby the more likely irrelevant information will be. To reduce the irrelevant information, for a given element $A_\mathrm{sa}^{ij}$ in self-attention map, the entropy gradient can be calculated as 
\begin{equation}
\frac{\mathrm{d} E}{\mathrm{d} A_\mathrm{sa}^{ij}} =  - (1 + \mathrm{log}(A_\mathrm{sa}^{ij})).
\label{eq:gradient_y}
\end{equation}
Finally, we calculate the entropy-reduced self-attention map by updating the original self-attention map along negative gradient direction as 
\begin{equation}
A_{\mathrm{sa}}^{ij} =  A_{\mathrm{sa}}^{ij} + \lambda(1+\mathrm{log}(A_{\mathrm{sa}}^{ij})),
\label{eq:ent_reduced_self_att}
\end{equation}
where $\lambda$ represents the updating factor.

\begin{figure}[t]
\centering
\footnotesize
\includegraphics[width=\linewidth]{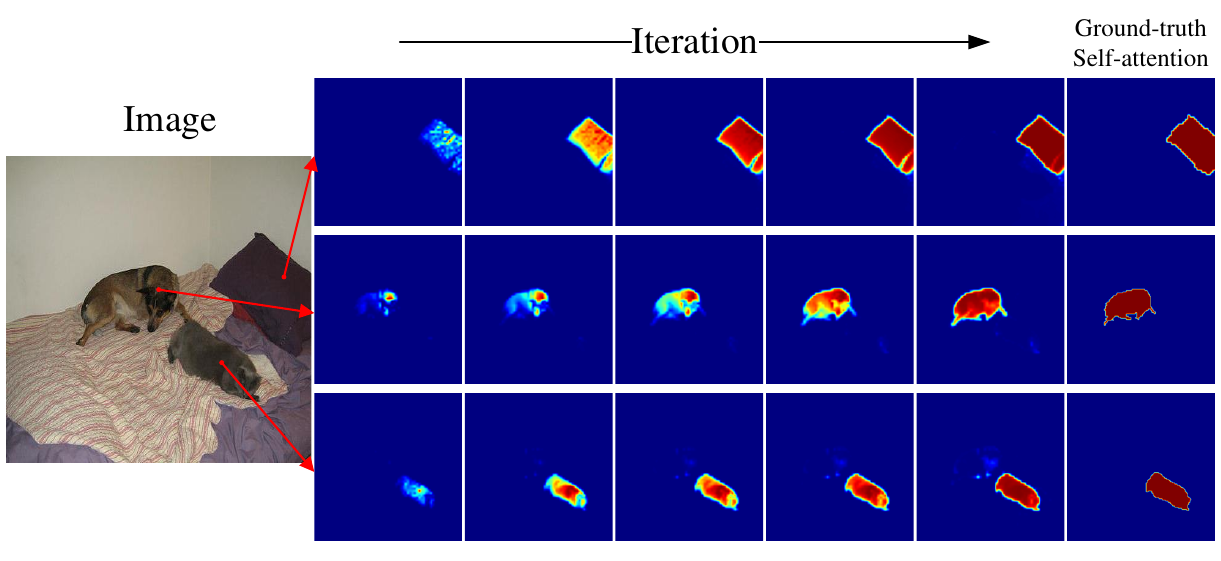}
\caption{\textbf{Refined self-attention map at selected points.} Our iterative refinement with entropy-reduced self-attention can be viewed as to generate the better refined self-attention map. After multiple iterations, the refined self-attention map becomes more similar to the ground-truth self-attention map at corresponding red point. The refined self-attention map can better refine the cross-attention map.}
\label{fig:gt_sa}
\end{figure}
\textbf{Analysis on effectiveness.} The reason why our iterative refinement with entropy-reduced self-attention module leads to better segmentation results can be explained from  two perspectives. (i) The entropy-reduced operation makes  the  value of $A_{\mathrm{sa}}^{ij}$ near to one becomes larger, and makes the value of $A_{\mathrm{sa}}^{ij}$ near to zero becomes smaller. As a result, the entropy-reduced operation can suppress the weak responses within self-attention map corresponding to irrelevant global information, and does not reduce the strong responses corresponding to the  local region. With this entropy-reduced operation, our iterative refinement can progressively enhance the local object region within cross-attention map without highlighting irrelevant regions during multiple iterations. (ii) Our iterative refinement can be viewed as to generate the better refined self-attention map, which is then used to refine cross-attention map once. Specifically, our iterative refinement can be approximately written as $A_{\mathrm{ca}}^{n} = (A_{\mathrm{sa}}^{\mathrm{ent}})^n \ast A_{\mathrm{ca}}$, where $(A_{\mathrm{sa}}^{\mathrm{ent}})^n$ can be seen as a refined self-attention map after $n$ iterations. We observe that the refined self-attention map $A_{\mathrm{ca}}^{n}$ becomes  clean and closer to the ground-truth self-attention map after multiple iterations.  Fig. \ref{fig:gt_sa} shows some examples of self-attention map at selected points. Therefore, compared to original self-attention map, the refined self-attention map can better improve cross-attention map.

\begin{figure}[t]
\centering
\setlength{\belowcaptionskip}{-10pt}
\includegraphics[width=\linewidth]{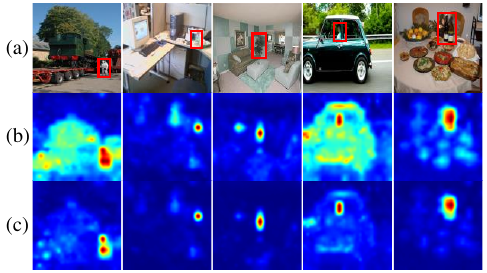}
\caption{\textbf{Comparison of cross-attention maps} before and after Cat-Cross module. Compared to the original cross-attention map (b), the refined cross-attention map (c) is more clean, and has strong response around corresponding objects in red bounding-box.}
\label{fig:11compare_cat}
\end{figure}

\noindent\textbf{Category-enhanced cross-attention module.} The accurate cross-attention map presents a good initial state for iterative refinement. Ideally, if we have a perfect cross-attention map, we do not require self-attention refinement. 
As show in Fig. \ref{fig:11compare_cat}(b), the original cross-attention maps usually have the strong responses on irrelevant regions, such as other categories or background. To improve the quality of cross-attention map fed to the refinement, we introduce a simple category-enhanced cross-attention (Cat-Cross) module, which makes the attention focus more on the text features belonging to the given categories by a factor of $\gamma$. Specifically, for a set of given categories $\mathcal{C}$ for segmentation, we  generate a weighting vector $W \in \mathcal{R} ^{T \times 1}$ as 
\begin{equation}
W[j] = 
\begin{cases}
  & \gamma,~~\text{ if } j\in \mathcal{C}, \\
  & 1,~~\text{ if } j\notin \mathcal{C},
\end{cases}
\label{eq:meaningful_mask}
\end{equation}
where  $\gamma$ is larger than $1$. With the weighting factor $W$, we generate weighted cross-attention map as
\begin{equation}
A_\mathrm{ca} = \mathrm{Softmax}(\frac{Q(W \cdot K)^{\mathrm{T}}}{\sqrt{d}}),
\label{eq:enhanced_attn}
\end{equation}
where $\cdot$ means element-wise multiplication, the query $Q$ represents visual features, and the key $K$ represents the text embeddings.

\begin{algorithm}[t]
\caption{Pipeline of our proposed iSeg} 
\label{alg:iterative_refinement}
\begin{algorithmic}[1]
\STATE \textbf{Input:}  \\ 
\STATE ~~~~$\gamma$: weighting factor of Cat-Cross Module  \\ 
\STATE ~~~~$\lambda$: updating factor of Ent-Self Module  \\ 
\STATE ~~~~$N$: iteration number\\
\STATE ~~~~$i$: category index\\
\STATE \textbf{Output:} \\ 
\STATE ~~~~$M_{\mathrm{p}}$: predicted segmentation mask
\STATE Generate latent features $z$ and text embeddings $\varepsilon$ of the image and text prompt, respectively
\STATE Compute category-enhanced text embeddings $\varepsilon_{\mathrm{en}}$ $\leftarrow$ Cat-Cross($\varepsilon$, $i$, $\gamma$)
\STATE Extract and select $A_{\mathrm{sa}}$ and $A_\mathrm{ca}$ from denoising UNet
\STATE Compute entropy-reduced self-attention $A_\mathrm{sa}^\mathrm{ent}$ $\leftarrow$ Ent-Self($A_{\mathrm{sa}}$, $\lambda$)
\STATE Set $n = 1$ and $A_\mathrm{ca}^\mathrm{0} = A_\mathrm{ca}$[i] at index  $i$ of text prompt
\REPEAT
    \STATE Normalize the  cross-attention map $A_\mathrm{ca}^\mathrm{n-1}$ 
    \STATE Update the refined cross-attention map $A_\mathrm{ca}^n \leftarrow A_\mathrm{sa}^\mathrm{ent} \ast A_\mathrm{ca}^{n-1}$
    \STATE Update $n \leftarrow n + 1$
\UNTIL{$n$ is larger than $N$}
\STATE Normalize the refined cross-attention map $A_\mathrm{ca}^N$
\STATE Generate the segmentation mask $M_{\mathrm{p}}$ by binarizing  $A_\mathrm{ca}^N$
\STATE \textbf{return} $M_{\mathrm{p}}$
\end{algorithmic}
\end{algorithm}

Algorithm \ref{alg:iterative_refinement} provides the pseudo code about how to refine the cross-attention by entropy-reduced self-attention iteratively using our proposed iSeg.

\section{Experiments}
\label{sec:exp}

In this section, to validate the effectiveness and superiority of our method, we perform the experiments on weakly-supervised semantic segmentation, open-vocabulary semantic segmentation, unsupervised segmentation, and mask generation for synthetic dataset.  Following the prior works \cite{Tian_2023_DiffSeg,Wang_2023_DiffSegmenter}, we mainly use pixel accuracy (ACC) and mean intersection over union (mIoU) to evaluate segmentation performance.

\begin{table*}[t]
\renewcommand{\arraystretch}{1.1}
\centering
\footnotesize
\caption{\textbf{Comparison of pseudo mask generation with weakly-supervised semantic segmentation approaches.} We report the mIoU results on PASCAL VOC 2012  and MS COCO training sets. Our proposed method outperforms various training-based and training-free approaches.}
\label{tab:sota_weakly}
\setlength{\tabcolsep}{4.5mm}{
\begin{tabular}{l|l|cc|cc}
\toprule
Type                          & Method                                     & Publication      & Training                                    & VOC                                    & COCO  \\
\midrule
\multirow{4}{*}{CNN-based}         & IRN \cite{Ahn_2019_IRN}                      & CVPR2019         & $\usym{1F5F8}$                              & 66.5                                   & 42.4                                  \\
                                   & AdvCAM \cite{Lee_2021_AdvCAM}                & CVPR2021         & $\usym{1F5F8}$                              & 55.6                                   & 35.8                                  \\
                                   & BAS \cite{Zhai_2024_BAS}                     & IJCV2023         & $\usym{1F5F8}$                              & 57.7                                   & 36.9                                   \\
                                   & HSC \cite{Wu_2023_HSC}                       & IJCAI2023        & $\usym{1F5F8}$                              & 71.8                                   & -                                      \\
\midrule
\multirow{4}{*}{Transformer-based} & MCTformer \cite{Xu_2022_MCTFormer}           & CVPR2022         & $\usym{1F5F8}$                              & 61.7                                   & -                                      \\
                                   & MCTformer+ \cite{Xu_2023_MCTFormer+}         & arXiv2023        & $\usym{1F5F8}$                              & 68.8                                   & -                                      \\
                                   & ToCo \cite{Ru_2023_ToCo}                     & CVPR2023         & $\usym{1F5F8}$                              & 72.2                                   & -                                      \\
                                   & WeakTr \cite{Zhu_2023_WeakTr}                & arXiv2023        & $\usym{1F5F8}$                              & 66.2                                   & -                                      \\
\midrule
\multirow{2}{*}{CLIP-based}        & CLIMS \cite{Xie_2022_Clims}                  & CVPR2022         & $\usym{1F5F8}$                              & 56.6                                   & -                                      \\
                                   & CLIP-ES \cite{Lin_2023_CLIPES}                 & CVPR2023         & $\usym{2717}$                               & 70.8                                   & 39.7                                   \\
\midrule
                                   & DiffSegmenter \cite{Wang_2023_DiffSegmenter}     & arXiv2023        & $\usym{2717}$                               & 70.5                                   & -                                      \\
                                   & T2M \cite{Xiao_2023_T2M}                    & arXiv2023        & $\usym{2717}$                               & 72.7                                   & 43.7                                   \\ 
\rowcolor{gray!15} \multirow{-3}{*}{\cellcolor{white}Diffusion-based}  & \textbf{iSeg (Ours)}                       & -                & $\usym{2717}$                               & \textbf{75.2}                          & \textbf{45.5}                          \\
\bottomrule
\end{tabular}}
\end{table*}

\subsection{Implementation Details}
We employ the pre-trained stable diffusion V2.1 as the base model, and feed the input image into the pre-trained model for feature extraction without any training on segmentation dataset. The input image is resized to 512$\times$512 pixels. All segmentation tasks are evaluated on a single NVIDIA RTX 3090 GPU with 24G memory. During inference, only 5G memory is required when using half precision with a batch size of 1, resulting in an approximate inference time of 0.16 seconds per image. We present the implementation details on various segmentation tasks below.

\noindent \textbf{Weakly-supervised semantic segmentation.} 
In this task, one  key is to generate pseudo pixel-level masks for  training set. 
We compare our training-free method with existing approaches for pixel-level mask generation on two public datasets, including PASCAL VOC 2012 \cite{Everingham_2010_PASCAL} and MS COCO 2014 \cite{Lin_2014_COCO}. 
Specifically, we use
the categories existing in the image to generate a text template such as \textit{`A photograph of {\rm[CLS A]} and {\rm[CLS B]} and other objects and background'} and add the background categories, similar to CLIPES \cite{Lin_2023_CLIPES}.

\begin{table*}[t]
\renewcommand{\arraystretch}{1.1}
\centering
\footnotesize
\caption{\textbf{Comparison with open-vocabulary segmentation approaches.} We reports the mIoU results on PASCAL VOC 2012 validation set, PASCAL-VOC Context  validation set, and MS COCO-Object validation set. Our proposed method achieves the promising performance.}
\label{tab:sota_open}
\setlength{\tabcolsep}{3.5mm}{
\begin{tabular}{l|c|cc|ccc}
\toprule
Type                          & Method                                     & Publication              & Training                    & VOC                       & Context                      & Object                  \\
\midrule
\multirow{7}{*}{CLIP-based}        & ReCo    \cite{Shin_2022_RECO}                & NeurIPS2022              & $\usym{1F5F8}$                       & 25.1                                      & 19.9                                      & 15.7                                   \\
                                   & MaskCLIP \cite{Zhou_2022_MaskCLIP}           & ECCV2022                 & $\usym{1F5F8}$                       & 38.8                                      & 23.6                                      & 20.6                                   \\
                                   & SegCLIP \cite{Luo_2023_SegCLIP}              & ICML2023                 & $\usym{1F5F8}$                       & 52.6                                      & 24.7                                      & 26.5                                   \\
                                   & CLIPpy \cite{Kanchana_2022_CLIPpy}           & ICCV2023                 & $\usym{1F5F8}$                       & 52.2                                      & -                                         & 32.0                                   \\
                                   & ViewCo \cite{Ren_2023_VIEWCO}                & ICLR2023                 & $\usym{1F5F8}$                       & 52.4                                      & 23.0                                      & 23.5                                   \\
                                   & OVSegmenter \cite{Xu_2023_OVSegmenter}              & CVPR2023                 & $\usym{1F5F8}$                        & 53.8                                      &       20.4                                   &     25.1                                  \\                                   
                                   & TCL \cite{cha2022tcl}              & CVPR2023                 & $\usym{1F5F8}$                        & 51.2                                      & 24.3                                         & 30.4                                      \\
                                   & TagCLIP \cite{Lin_2024_TagCLIP}              & AAAI2024                 & $\usym{2717}$                        & 64.8                                      & -                                         & -                                      \\                            
                                   & CaR \cite{sun_2024_car}              & CVPR2024                 & $\usym{2717}$                        & 67.6                                      & 30.5                                         & 36.6                                      \\
\midrule                                
 SAM-based     & SAM-CLIP  \cite{Wang_2024_SAMCLIP}                    & CVPRW2024          & $\usym{1F5F8}$                        & 60.6                                      & 29.2                            & 31.5                         \\                                   
\midrule
                                   & OVDiff \cite{Karazija_2023_OVDiff}           & ECCV2024                & $\usym{2717}$                        & 67.1                                & 30.1                                      & 34.8                                   \\
                                   & DiffSegmenter \cite{Wang_2023_DiffSegmenter}     & arXiv2023                & $\usym{2717}$                        & 60.1                                      & 27.5                                      & 37.9                                   \\ 
\rowcolor{gray!15} \multirow{-3}{*}{\cellcolor{white}Diffusion-based}        & \textbf{iSeg (Ours)}                       & -          & $\usym{2717}$                        & \textbf{68.2}                                      & \textbf{30.9}                             & \textbf{38.4}                          \\
\bottomrule
\end{tabular}}
\end{table*}

\noindent \textbf{Open-vocabulary semantic segmentation.} This task aims to perform segmentation of arbitrary categories. We compare our training-free method with existing 
approaches on three public datasets, including PASCAL VOC 2012 (VOC), PASCAL-Context (Context), and MS COCO Object (Object).  To perform open-vocabulary semantic segmentation, we employ the pre-trained CLIP to generate image-level category labels as text prompts following 
TagCLIP \cite{Lin_2024_TagCLIP}. Afterwards, we keep the similar settings as weakly-supervised semantic segmentation task to obtain the final segmentation masks of different categories.

\noindent \textbf{Unsupervised segmentation.} This task groups pixels into different objects without using any annotations. Here we compare our  method with other unsupervised segmentation approaches on two public datasets, including COCO-Stuff-27 \cite{Caesar_2018_COCOStuff} and Cityscapes \cite{Cordts_2016_Cityscapes}. We integrate our proposed method into the training-free approach DiffSeg \cite{Tian_2023_DiffSeg} as a post-processing. Specifically, we treat the segmentation results of DiffSeg as initial cross-attention maps, and employ our proposed entropy-reduced self-attention module to perform iterative refinement, while the other settings are consistent with DiffSeg.

\noindent \textbf{Mask generation for synthetic dataset.} Recently,  researchers have explored generating synthetic dataset for semantic segmentation. In this task, it is important to generate precise  mask labels for synthetic images. DiffuMask \cite{Wu_2023DiffuMask} utilizes cross-attention maps in diffusion model to generate pseudo masks for synthetic images, which are further refined by AffinityNet. Here, we can directly apply our iterative refinement to improve the initial pseudo masks generated by DiffuMask on synthetic dataset without requiring any additional refinements like AffinityNet.

\begin{table*}[t]
\renewcommand{\arraystretch}{1.1}
\centering
\footnotesize
\caption{\textbf{Comparison with some unsupervised semantic segmentation approaches.} We report the results on Cityscapes and COCO-Stuff-27 validation sets. Our iSeg stably outperforms DiffSeg and other approaches on these two datasets in terms of mIoU and ACC.}
\label{tab:sota_unsupervised}
\setlength{\tabcolsep}{5.0mm}{
\begin{tabular}{l|cc|cc|cc}
\toprule
                    \multirow{2}{*}{Method}  &  \multirow{2}{*}{Publication} & \multirow{2}{*}{Training}  & \multicolumn{2}{c|}{Cityscapes}              & \multicolumn{2}{c}{COCO-Stuff-27}                 \\
                                                &                &                  & ACC                      & mIoU                      & ACC                     & mIoU                    \\
                    \midrule
                    MDC \cite{Caron_2018_MDC}      &     ECCV2018          & $\usym{1F5F8}$          & 40.7                     & 7.1                       & 32.3                    & 9.8                      \\
                    IIC \cite{Ji_2019_IIC}          &    ICCV2019    & $\usym{1F5F8}$          & 47.9                     & 6.4                       & 21.8                    & 6.7                      \\
                    PICLE \cite{Cho_2021_PICLE}     &    CVPR2021        & $\usym{1F5F8}$          & 65.5                     & 12.3                      & 48.1                    & 13.8                     \\
                    STEGO \cite{Hamilton_2022_STEGO}   &    ICLR2022         & $\usym{1F5F8}$          & 73.2                     & 21.0                      & 56.9                    & 28.2                     \\                  
                    MaskCLIP \cite{Zhou_2022_MaskCLIP}    &    ECCV2022     & $\usym{1F5F8}$          & 35.9                     & 10.0                      & 32.2                    & 19.6                     \\
                    RoCo \cite{Shin_2022_RECO}      &    NeurIPS2022   & $\usym{1F5F8}$          & 74.6                     & 19.3                      & 46.1                    & 26.3                     \\
                    ACSeg \cite{Li_2023_ACSeg}   &    CVPR2023         & $\usym{1F5F8}$          & -                    &   -                    & -                    & 28.1                     \\  
                    \midrule
                    DiffSeg \cite{Tian_2023_DiffSeg}  &     CVPR2024       & $\usym{2717}$           & 76.0                     & 21.2                      & 72.5                    & 43.6                     \\

\rowcolor{gray!15}  \textbf{iSeg (Ours)}     &     -   & $\usym{2717}$           & \textbf{78.7}            & \textbf{25.0}             & \textbf{74.5}           & \textbf{45.2}      \\
\bottomrule
\end{tabular}}
\end{table*}

\subsection{State-of-the-art comparison}
\noindent\textbf{Weakly supervised semantic segmentation.} 
Table \ref{tab:sota_weakly} compares our proposed training-free method iSeg and some weakly-supervised methods, including CNN-based, transformer-based, CLIP-based, and diffusion-based approaches, on both PASCAL VOC \cite{Everingham_2010_PASCAL} and MS COCO \cite{Lin_2014_COCO} training sets. Here, we  compare the quality of generated pseudo masks of different methods on training set. Our proposed iSeg achieves the mIoU scores of 75.2\% and 45.5\% on PASCAL VOC and MS COCO, respectively. Compared to diffusion-based training-free methods DiffSegmenter \cite{Wang_2023_DiffSegmenter} and T2M \cite{Xiao_2023_T2M}, our iSeg has 4.7\% and 2.5\% improvements in terms of mIoU on PASCAL VOC. Compared to CLIP-based method CLIP-ES \cite{Lin_2023_CLIPES}, our proposed method iSeg has 4.4\% and 5.8\% improvements in terms of mIoU on PASCAL VOC and MS COCO. In addition, our proposed iSeg  outperforms some training-based approaches. For instance, compared to training-based methods ToCo \cite{Ru_2023_ToCo} and WeakTr \cite{Zhu_2023_WeakTr}, our proposed iSeg has 3.0\% and 9.0\% improvements in terms of mIoU on PASCAL VOC.

\noindent\textbf{Open-vocabulary semantic segmentation.} Table \ref{tab:sota_open} compares our proposed method iSeg with some open-vocabulary semantic segmentation methods, including CLIP-based, SAM-based and diffusion-based approaches, on PASCAL VOC validation set, PASCAL Context validation set \cite{Mottaghi_2014_Context}, and MS COCO Object validation set. Our proposed iSeg achieves the mIoU scores of 68.2\%, 30.9\% and 38.4\% on these three datasets, respectively. Compared to training-free DiffSegmenter \cite{Wang_2023_DiffSegmenter},  our proposed iSeg has the absolute gain of $8.1\%$, $3.4\%$, and $0.5\%$ on these three datasets, respectively.  Compared to training-free OVDiff \cite{Karazija_2023_OVDiff},  our proposed iSeg has the absolute gain of $1.1\%$, $0.8\%$, and $3.6\%$, respectively.  In addition, our diffusion-based iSeg outperforms CLIP-based approaches, such as SegCLIP \cite{Luo_2023_SegCLIP}, ViewCo \cite{Ren_2023_VIEWCO}, and TagCLIP \cite{Lin_2024_TagCLIP}. For instance, our iSeg outperforms ViewCo \cite{Ren_2023_VIEWCO} by the absolute gain of $15.8\%$, $7.9\%$, and $14.9\%$ on these three datasets. The CLIP-based CaR \cite{sun_2024_car}  achieves comparable performance with our proposed method iSeg. However, CaR suffers from slow inference speed ($>$1s per image) due to heavy  feature extraction at multiple iterations. Moreover, our training-free iSeg is superior to training-based SAM-CLIP \cite{Wang_2024_SAMCLIP}, which outperforms SAM-CLIP \cite{Ren_2023_VIEWCO} by the absolute gain of $7.6\%$, $1.7\%$, and $6.9\%$ on three datasets, respectively.

\begin{table*}[t]
\renewcommand{\arraystretch}{1.1}
\centering
\footnotesize
\caption{\textbf{Impact of integrating  the two  modules.} We reports the results on various tasks and datasets using the metric of mIoU. The baseline in first row simply refines cross-attention map by self-attention map once, similar to existing methods. Our proposed modules, including entropy-reduced self-attention (Ent-Self) and category-enhanced cross-attention (Cat-Cross), can significantly improve the performance of the baseline.}
\setlength{\tabcolsep}{2.8mm}{
\begin{tabular}{cc|cc|ccc|cc}
\toprule
\multirow{2}{*}{Ent-Self}              &   \multirow{2}{*}{Cat-Cross}               & \multicolumn{2}{c|}{Weakly-supervised}         & \multicolumn{3}{c|}{Open-vocabulary}  & \multicolumn{2}{c}{Unsupervised}                                             \\
              &                 & VOC        & COCO  & VOC  & Context  & Object  & Cityscapes   & COCO-Stuff                                          \\
\midrule
 $\usym{2717}$         & $\usym{2717}$             & 68.2  & 40.1    & 63.7	& 26.4  & 36.6  & 22.8	& 44.4 \\
 $\usym{1F5F8}$        & $\usym{2717}$             & 72.0  & 42.5	& 67.1	& 28.2	& 37.5	& 25.0	& 45.2\\                     
\rowcolor{gray!15}   $\usym{1F5F8}$        & $\usym{1F5F8}$            & \textbf{75.2} & \textbf{45.5}	& \textbf{68.2}	& \textbf{30.9}	& \textbf{38.4}	& N/A	&N/A\\

\bottomrule
\label{ablation:modules}
\end{tabular}}
\end{table*}

\noindent\textbf{Unsupervised segmentation.} 
Table \ref{tab:sota_unsupervised} compares our  method with some unsupervised segmentation methods on Cityscapes \cite{Cordts_2016_Cityscapes} validation set and COCO-Stuff-27  \cite{Caesar_2018_COCOStuff} validation set.
We employ our iterative refinement as a post-processing, and integrate it into unsupervised DiffSeg \cite{Tian_2023_DiffSeg} to improve segmentation results. Our proposed iSeg  presents stable improvement on both Cityscapes and COCO-Stuff-27 validation sets. For instance, our iSeg outperforms DiffSeg by a large gain of $2.7\%$ in terms of ACC and $3.8\%$ in terms of mIoU on Cityscapes.  In addition, our proposed method outperforms some training-based methods.  For instance, our iSeg outperforms MaskCLIP \cite{Zhou_2022_MaskCLIP} and RoCo \cite{Shin_2022_RECO} by  $15.0\%$ and 5.7\% in terms of ACC on Cityscapes and COCO-Stuff-27, respectively.

\begin{figure}[t]
\centering
\includegraphics[width=\linewidth]{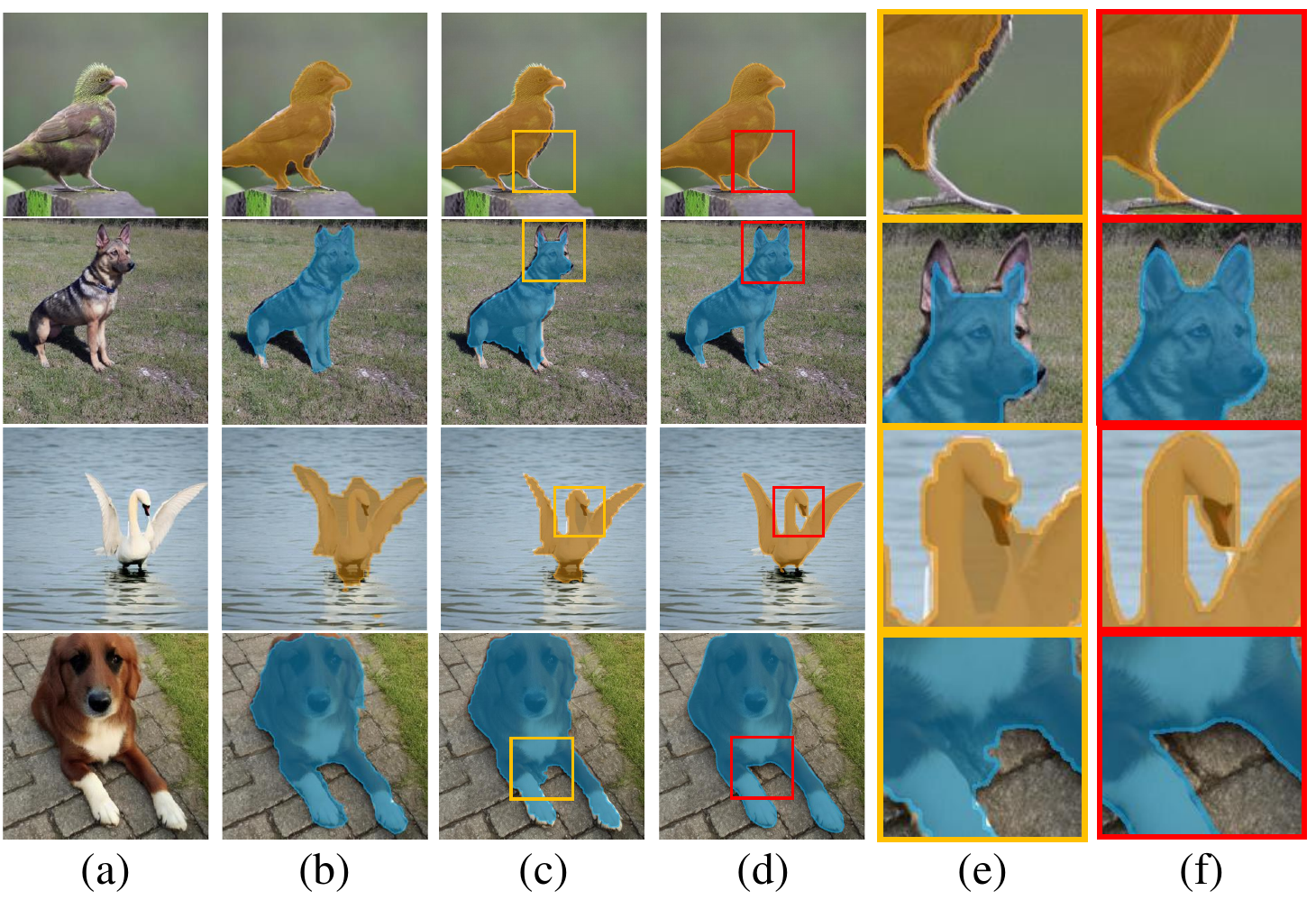}
\caption{\textbf{Mask generation on synthetic dataset.} For synthetic images in column (a), DiffuMask first generates coarse pseudo masks using cross-attention maps in column (b) and refines them with AffinityNet in column (c). Instead of using AffinityNet, we employ our  method to refine  cross-attention maps in column (d). Compared to AffinityNet, our  method can generate more accurate masks of different categories for these synthetic images, zoomed in (e,f).}
\label{fig:06compare}
\end{figure}

\noindent\textbf{Mask generation on synthetic dataset.} Fig. \ref{fig:06compare} gives some qualitative results of our proposed iSeg and DiffuMask \cite{Wu_2023DiffuMask}. The synthetic images (a) are  from DiffuMask \cite{Wu_2023DiffuMask}. For mask generation, DiffuMask first generates coarse pseudo masks (b) using  cross-attention map, and then trains an AffinityNet to generate more precise pseudo masks (c). Compared to using AffinityNet, our proposed iSeg has more accurate segmentation results (d) without using any training processing. For better visualization, we zoom the local segmentation results of AffinityNet and our iSeg in (e) and (f). For instance, our proposed iSeg provides more segmentation about dog outline in second row.

\subsection{Ablation study}

Here we perform the ablation study to demonstrate the effectiveness of our proposed modules.

\noindent\textbf{Impact of integrating two proposed modules.} Table \ref{ablation:modules} shows the results on various datasets and tasks. To demonstrate the effectiveness of our iterative refinement, the baseline simply refines cross-attention map with self-attention map only once. When integrating  entropy-reduced self-attention (Ent-Self) module, it achieves the performance improvement on all the tasks and datasets. For instance, it presents 3.8\% improvement on VOC for weakly-supervised semantic segmentation,  3.4\% improvement on VOC for open-vocabulary semantic segmentation, and 2.2\% improvement on CityScapes for open-vocabulary semantic segmentation.
By further integrating both Ent-Self and Cat-Cross into the baseline, it has the best performance. For instance, it improves the baseline by 7.0\% on VOC for weakly-supervised semantic segmentation, and 4.5\% on both VOC and COCO Object for open-vocabulary semantic segmentation. 
It note that, our Cat-Cross module is not used for unsupervised segmentation task, because this unsupervised task does not requires category information.

\noindent\textbf{Hyper-parameter settings.} There are some hyper-parameters in our method, including the number $N$ of iterations, the updating factor $\lambda$ in our Ent-Self module, and the weighting factor $\gamma$ in our Cat-Cross module. We conduct the ablation study of hyper-parameter setting on VOC for weakly-supervised semantic segmentation. Table \ref{ablation:hyper}(a) presents the impact of $N$. With the increasing iterations, it has the performance improvement. When the number $N=10$, it has the best result. Table \ref{ablation:hyper}(b) shows the impact of updating factor $\lambda$. When the updating factor $\lambda=0.01$, it achieves the best performance. When the updating factor becomes larger or smaller, it has some performance drop.  Table \ref{ablation:hyper}(c) gives the impact of weighting factor $\gamma$. When the weighting factor $\gamma=1.6$, it has the best performance.

\begin{table}[t]
\renewcommand{\arraystretch}{1.1}
\centering
\footnotesize
\caption{\textbf{Hyper-parameter study} of our proposed modules. Here, we present the impact of different hyper-parameters in our iterative refinement framework, including the number $N$ of iterations (a), the updating factor $\lambda$ in our Ent-Self, and the weighting factor $\gamma$ in our Cat-Cross.}
\subfloat[Iteration]{
\setlength{\tabcolsep}{2.0mm}{
\begin{tabular}{c|ccccccc}
\toprule
$N$      & 1     & 2      & 4       & 6      & 8             & 10               & 12\\
\midrule
mIoU     & 71.0  & 72.9   & 74.5	& 75.0	 & 75.1          & \textbf{75.2}	& 74.9    \\
\bottomrule
\end{tabular}}}
\hfill
\subfloat[Updating factor]{
\setlength{\tabcolsep}{2.4mm}{
\begin{tabular}{c|cccccc}
\toprule
$\lambda$  & 0     & 0.001    & 0.005  & 0.01          & 0.05  & 0.1      \\
\midrule
mIoU       & 69.1      & 74.3	   & 75.0	& \textbf{75.2} & 74.6	& 74.1    \\
\bottomrule
\end{tabular}}}
\hfill
\subfloat[Weighting factor]{
\setlength{\tabcolsep}{2.5mm}{
\begin{tabular}{c|cccccc}
\toprule
$\gamma$  & ~~1~~     & 1.2     & 1.4   & 1.6                 & 1.8     & 2      \\
\midrule
mIoU      & 72.0  & 73.6	& 74.7	& \textbf{75.2}       & 75.2	& 74.9    \\
\bottomrule
\end{tabular}}}
\label{ablation:hyper}
\end{table}

\begin{figure}[t]
\centering
\includegraphics[width=\linewidth]{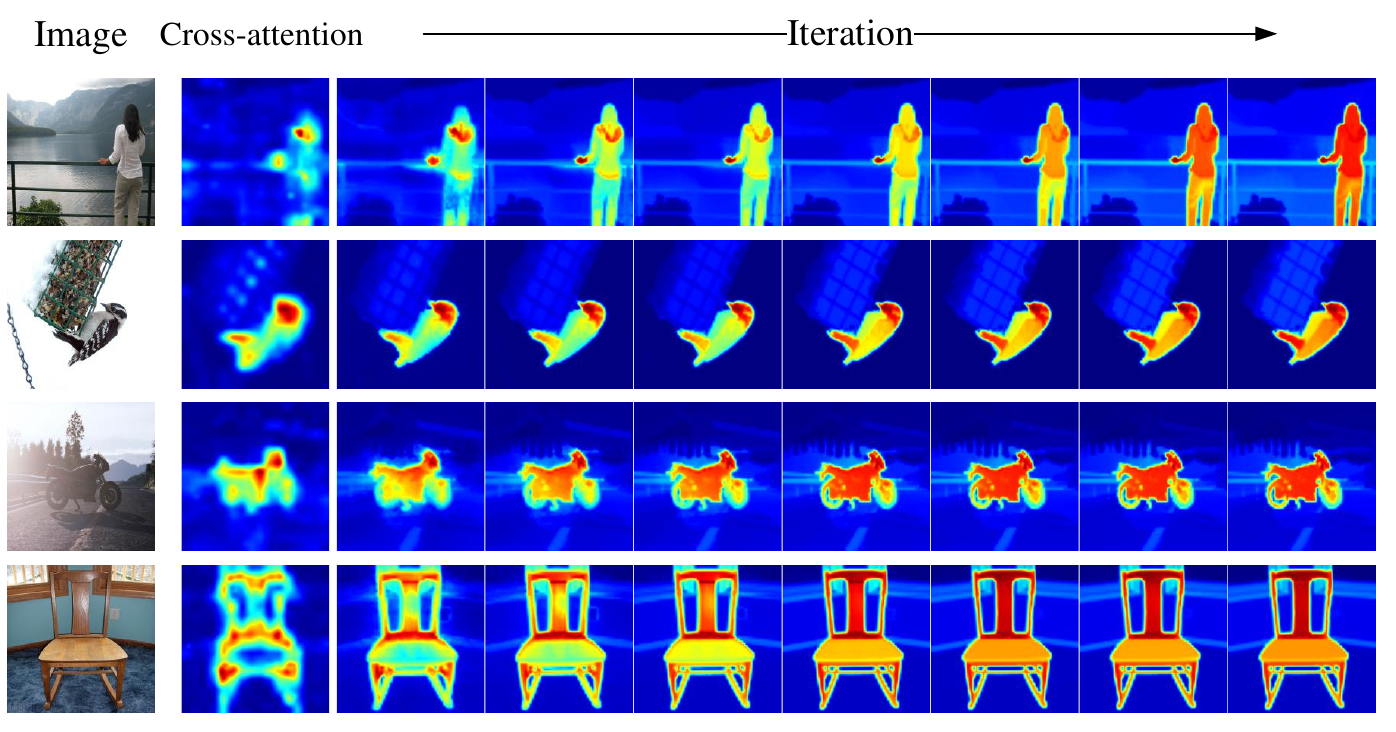}
\caption{\textbf{Refined cross-attention maps} of our proposed iSeg at different iterations. With the increment of iterations, our proposed method can provide more accurate response around the given objects.}
\label{fig:visual_iter}
\end{figure}

\begin{table}[t]
\renewcommand{\arraystretch}{1.1}
\centering
\footnotesize
\caption{\textbf{Impact of different designs} when using pre-trained stable diffusion model. We first present the impact of using cross-attention maps at different decoder layers. Then, we compare the impact of using self-attention maps at different decoder layers in (b). Finally, we show the impact of different time-steps in (c).}
\subfloat[Cross-attention map]{
\setlength{\tabcolsep}{6.5mm}{
\begin{tabular}{c|ccc}
\toprule
Level               & 16$\times$16            &  32$\times$32 &  Both\\
\midrule
mIoU        & 74.8	& 56.9	& \textbf{75.2}    \\
\bottomrule
\end{tabular}}}
\hfill
\subfloat[Self-attention map]{
\setlength{\tabcolsep}{7.5mm}{
\begin{tabular}{c|ccc}
\toprule
Layer       &  \#-3 &	\#-2	&\#-1     \\
\midrule
mIoU        & 68.5	& 71.1 	& \textbf{75.2}    \\
\bottomrule
\end{tabular}}}
\hfill
\subfloat[Time-step]{
\setlength{\tabcolsep}{2.8mm}{
\begin{tabular}{c|ccccc}
\toprule
Number          &   ~~~~1~~~~     & ~~~50~~~ & ~~100~~ & ~~150~~  & ~~200~~\\
\midrule
mIoU        & 73.2	& 74.6 &	\textbf{75.2}	& 74.5  & 74.3    \\
\bottomrule
\end{tabular}}}
\label{ablation:designs}
\end{table}

\noindent \textbf{Cross-attention map after different iterations.}  Fig. \ref{fig:visual_iter} visualizes some refined cross-attention maps of our proposed iSeg at different iterations. With the increment of iterations, our proposed iSeg can generate more accurate cross-attention maps of corresponding objects. For instance, after multiple iterations, our method can make cross-attention map contour more precise and uniform, which benefits the high-quality segmentation.

\begin{figure}[t]
\centering
\includegraphics[width=\linewidth]{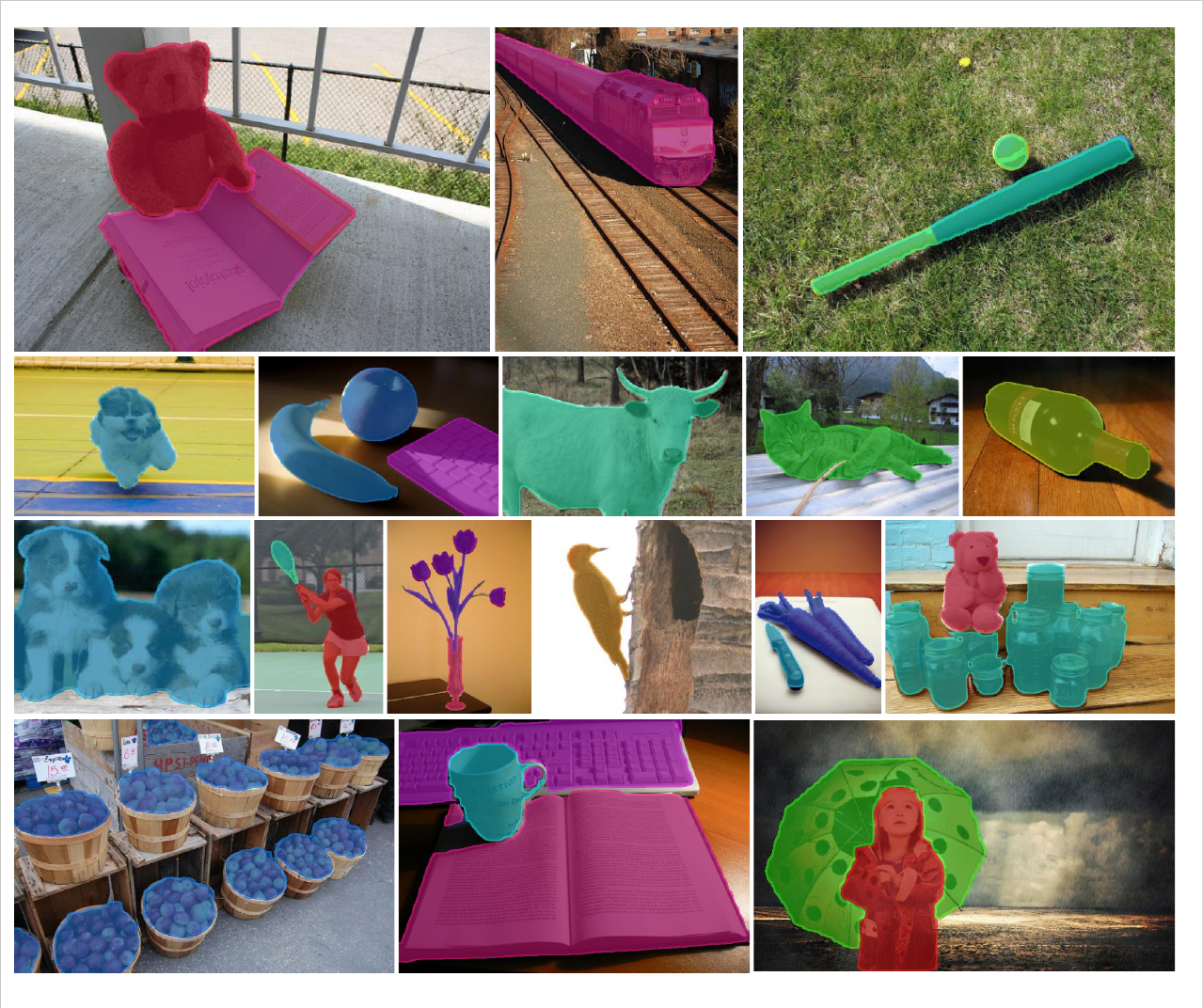}
\caption{\textbf{Segmentation results  of natural images} on PASCAL VOC 2012 and MS COCO 2014 datasets. Our proposed method is able to segment various categories under different situations.}
\label{fig:12visual}
\end{figure}

\noindent \textbf{Impact of different designs.}  Table \ref{ablation:designs} further gives the impact of different designs in our method,  including different layers (a,b) for cross-attention and self-attention maps, and different time-steps (c). In (a), we give the impact of using cross-attestation map at different levels. It demonstrates that, the fused cross-attention map of 16$\times$16 and 32$\times$32 pixels have the best result. In (b), we show the impact of different self-attention maps. \#-1 represents self-attention map at last  decoder layer. \#-3 and \#-2 fuse self-attention map of last decoder layer with that of third and second last decoder layer, respectively. It has the best result when using  self-attention map at last decoder layer.  In (c), we perform the experiments using different time-steps. We set time-step as 100 as our final settings, which has the best result.

\begin{figure}[t]
\centering
\includegraphics[width=\linewidth]{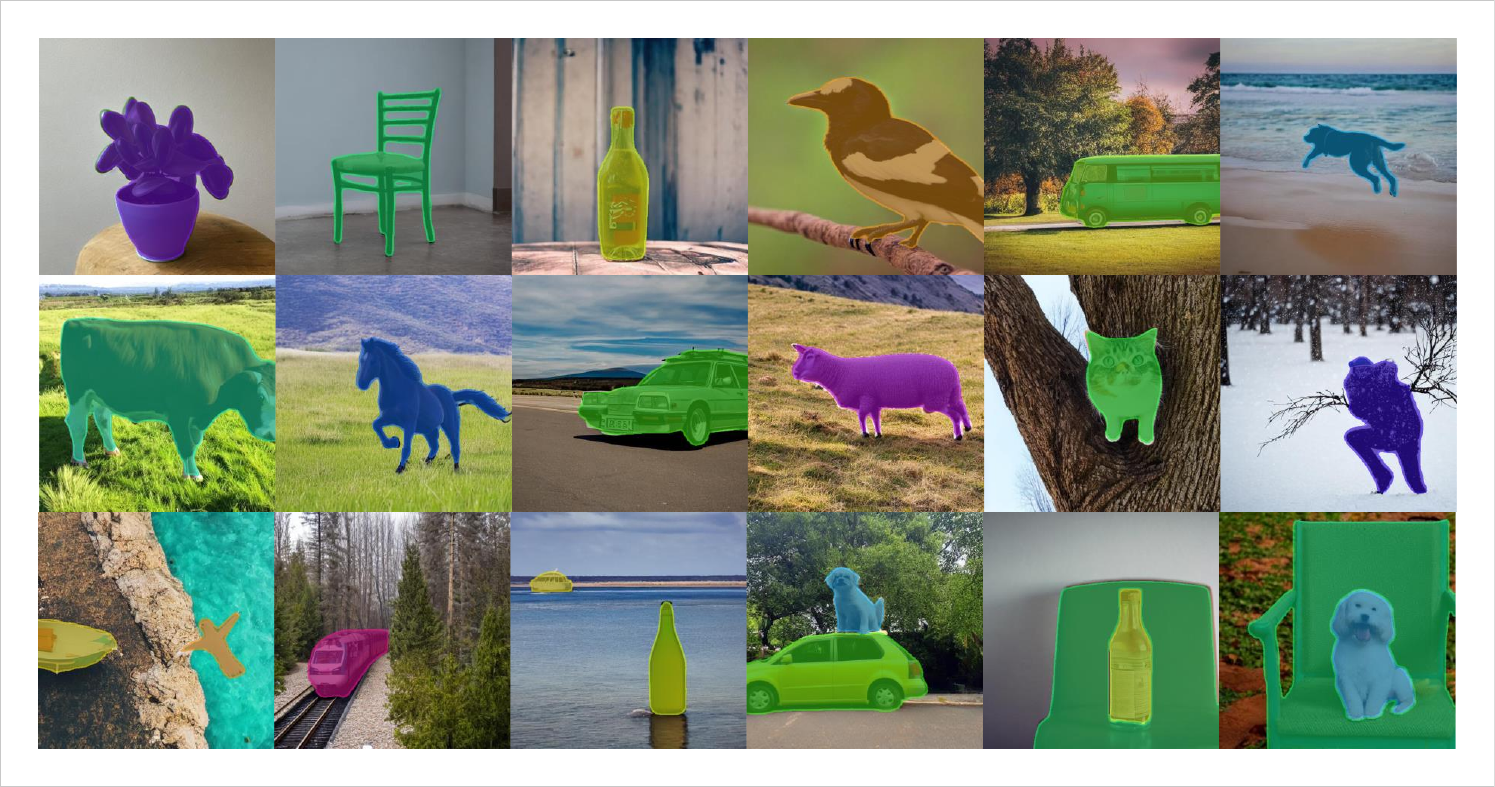}
\caption{\textbf{Segmentation results of synthetic images} with one or multiple categories. Our proposed method provide accurate segmentation masks for these synthetic objects.}
\label{fig:visual_synthesis}
\end{figure}

\begin{figure}[t]
\centering
\includegraphics[width=\linewidth]{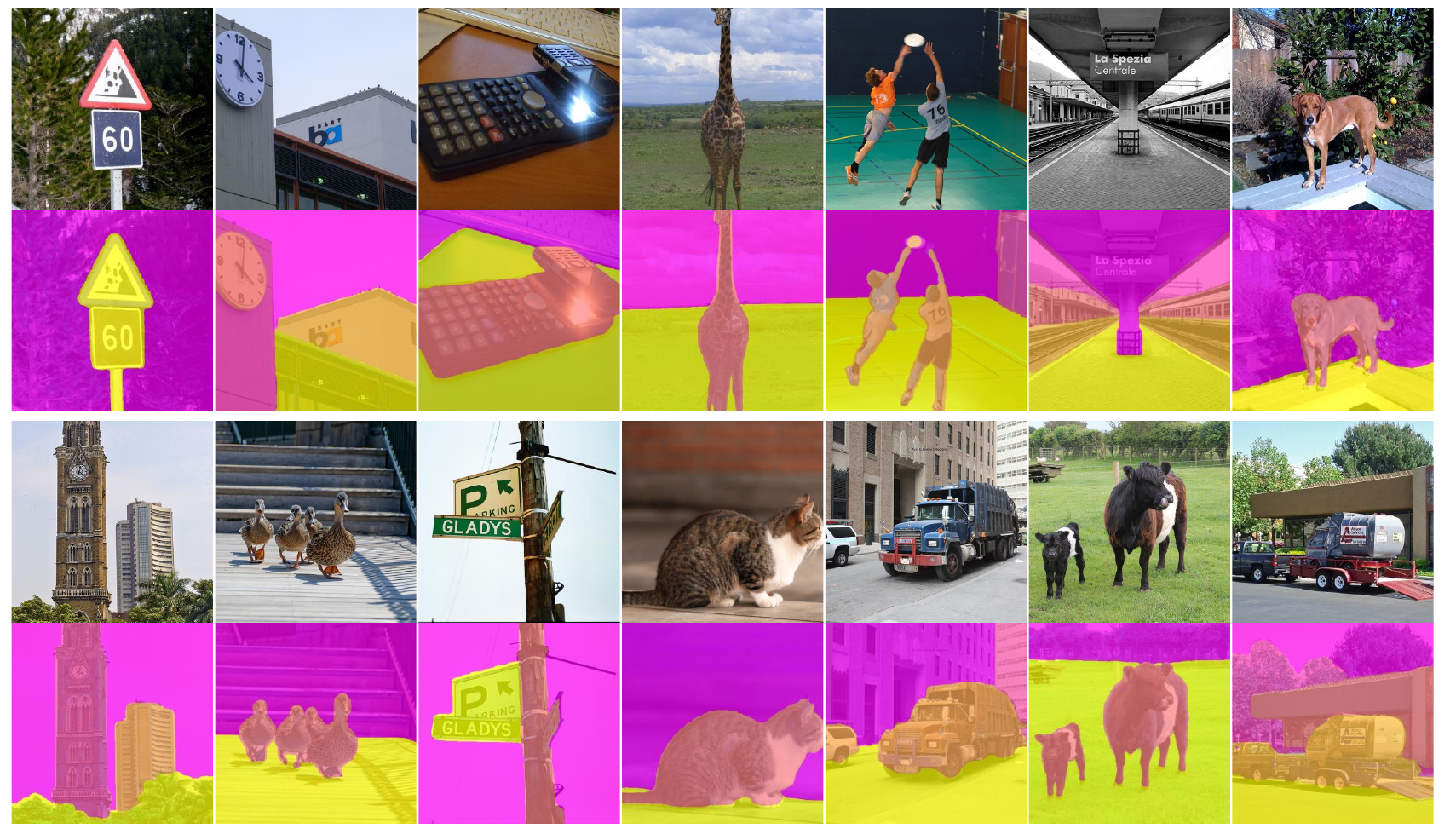}
\caption{\textbf{Unsupervised segmentation results} on COCO-Stuff-27 dataset. For given images, our proposed method can accurately group the pixels into different categories.}
\label{fig:13visual_unsupervise}
\end{figure}

\subsection{Qualitative Results}

We show segmentation results  of our proposed method on natural images, synthetic images, and unsupervised segmentation task.

\noindent \textbf{Results on  natural images.}  Fig. \ref{fig:12visual} presents some  segmentation results of our proposed iSeg. The images are from PASCAL VOC 2012 and MS COCO. It can be observed that our proposed method can generate accurate segmentation masks for various categories. For instance, our proposed iSeg accurately segments the tiny ball in third example, and  the adjacent person and umbrella in last example.

\noindent \textbf{Results on  synthetic images.}  Fig. \ref{fig:visual_synthesis} presents some  segmentation results of our proposed iSeg on synthetic images which are generated by image synthesis method DiffuMask \cite{Wu_2023DiffuMask}. Similar to natural images, our proposed method can accurately segment different categories for diverse synthetic images. For instance, our proposed iSeg generate accurate masks for different categories, such as potted plants, chairs, horses, birds, and cars.

\noindent \textbf{Results of unsupervised segmentation.}  Fig. \ref{fig:13visual_unsupervise} gives some unsupervised segmentation results of our  iSeg on COCO-Stuff-27 dataset. The first and third rows present the original images, while the second and fourth rows give the segmentation results. Our proposed method is able to accurately group the pixels into different categories.

\subsection{Extensions}
Here, we show that our proposed method   can support the segmentation for different kinds of cross-domain images, and perform the accurate segmentation using different interactions like point, line, and box.

\begin{figure}[t]
\centering
\includegraphics[width=0.99\linewidth]{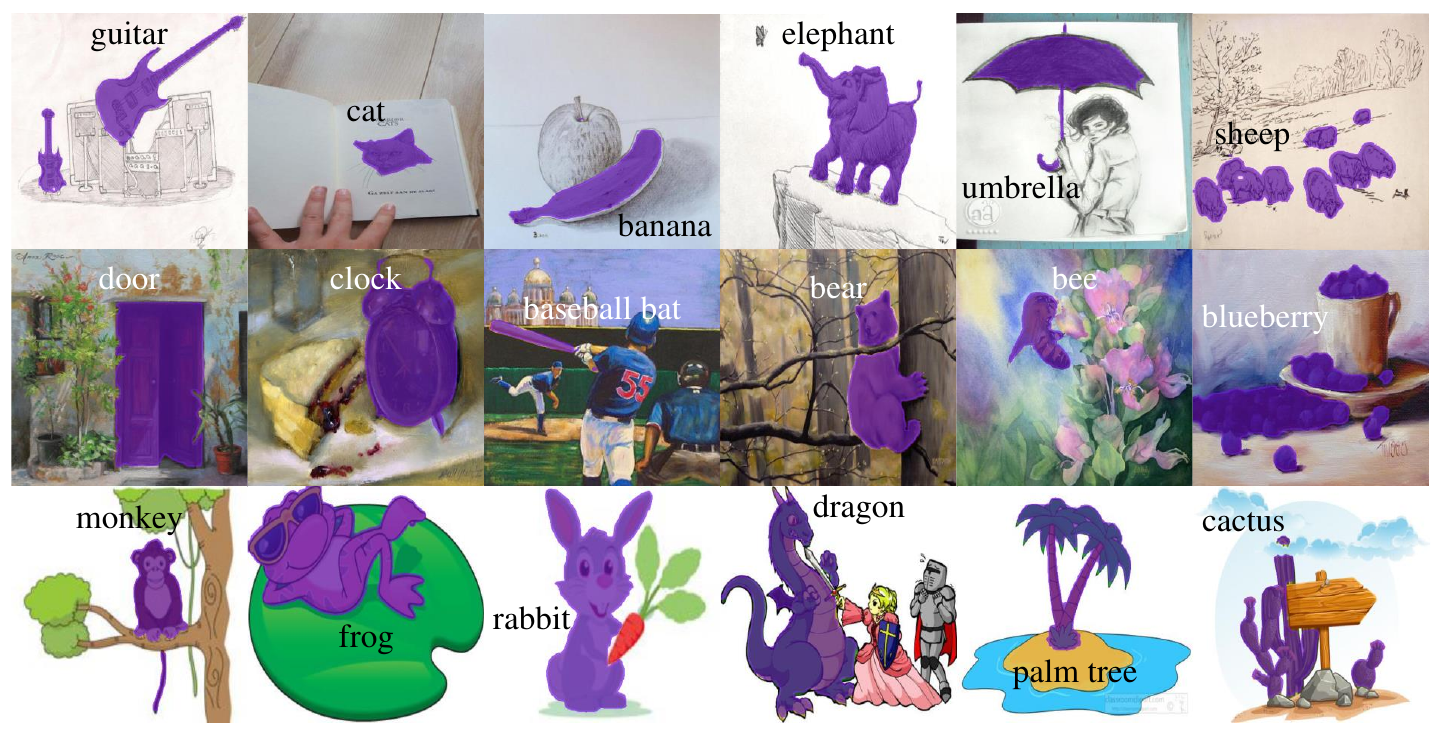}
\caption{\textbf{Cross-domain segmentation results} on DomainNet Sketch  (top), DomainNet Painting  (middle), and DomainNet Clipart  (bottom). Our method presents accurate segmentation results for these cross-domain images.}
\label{fig:visual_domain}
\end{figure}

\begin{figure}[t]
\centering
\includegraphics[width=\linewidth]{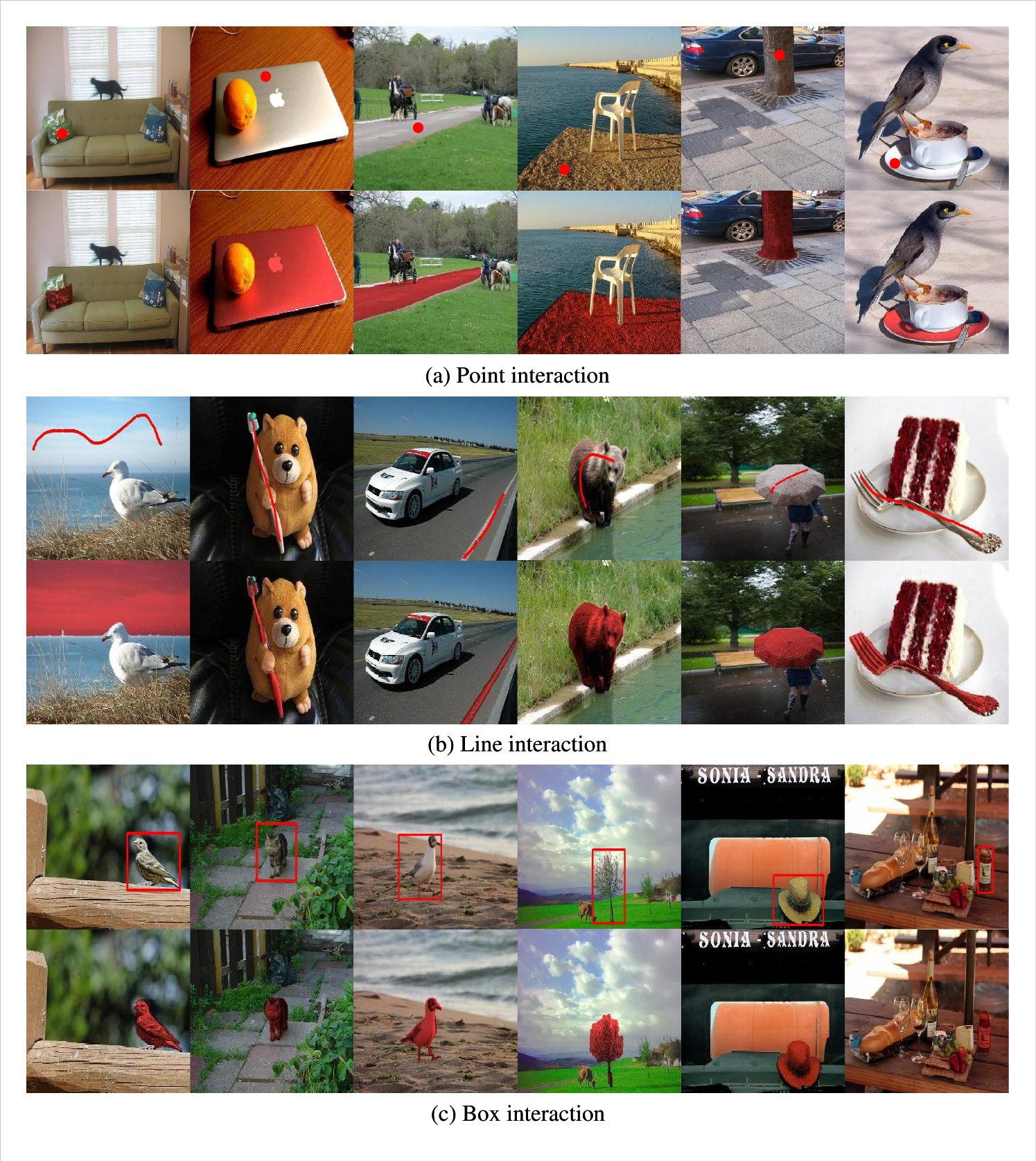}
\caption{\textbf{Segmentation results using different interactions}, including points, boxes, and lines. By multiple iterations, our method can accurately segment the objects according to different interactions.}
\label{fig:visual_interaction}
\end{figure}

\noindent\textbf{Different kinds of images.} Fig. \ref{fig:visual_domain} presents some kinds of image segmentation examples on DomainNet \cite{Peng_2019_DomainNet}.  The DomainNet includes three parts: DomainNet Sketch, DomainNet Painting, and DomainNet Clipart, which are significantly different to natural images and synthetic images. Our proposed method is able to accurately segment different kinds of objects including sketch, painting and clipart according to given categories. For instance, our method accurately segments the guitar in first sketch image, bear in fourth painting image, and cactus in last clipart image. 

\noindent\textbf{Different interactions.} Fig. \ref{fig:visual_interaction} presents more segmentation results using different interactions, such as points, lines, and boxes. The first row in each sub-figure shows the image and corresponding interaction. We first generate the initial cross-attention maps. Specifically, we  set the  pixels belonging to points and lines as foreground object, or set the pixels within boxes as foreground object.  By iterative refinement, our method can refine the initial cross-attention maps, and thus accurately segment the objects corresponding to points, lines, or boxes, as in second row. For instance, our proposed method accurately segments the selected laptop computer without mistaking the orange on the laptop as the laptop.

\section{Conclusion}
\label{sec:conclusion}

In this paper, we propose an iterative refinement framework for training-free segmentation using pre-trained stable diffusion model, named iSeg. Our  iSeg introduces two novel modules: entropy-reduced self-attention module and category-enhanced cross-attention module. The entropy-reduced self-attention module aims to suppress irrelevant global information within self-attention map, while the category-enhanced cross-attention module pays more attention to the features of given segmented categories. With these two proposed modules, our proposed method can generate accurate segmentation masks without any training on segmentation datasets. We perform the experiments on various datasets and  segmentation tasks to demonstrate the efficacy of our proposed method.

\noindent\textbf{Limitations and future work.}
Currently, we set the number of iterations in our iterative refinement  as a fixed number, which is not optimal. In fact, the number of iterations can be dynamically changed according to the image. In future, we will explore image-aware iterative refinement framework.


%





\ifCLASSOPTIONcaptionsoff
  \newpage
\fi



%


{\small
\bibliographystyle{ieee}
\bibliography{refs1}
}

%


\begin{IEEEbiography}[{\includegraphics[width=1in,height=1.25in,clip,keepaspectratio]{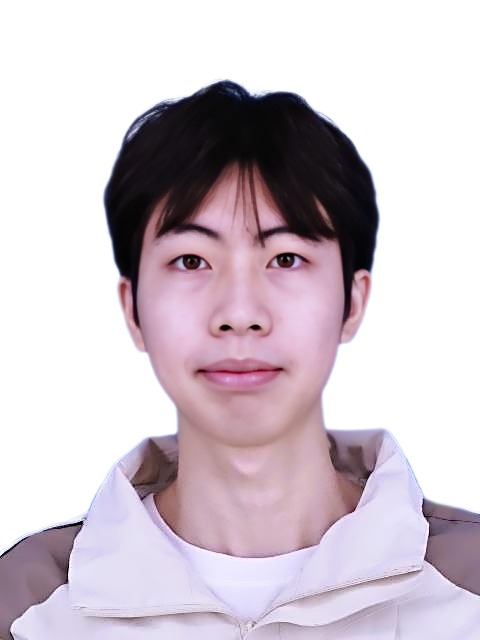}}]{Lin Sun} received the B.S. in electronic information engineering from the Shandong University, Weihai, China, in 2023. He is currently pursuing his master degree at the Tianjin University. His research interests include image segmentation and vision-language learning.
\end{IEEEbiography}

\begin{IEEEbiography}[{\includegraphics[width=1in,height=1.25in,clip,keepaspectratio]{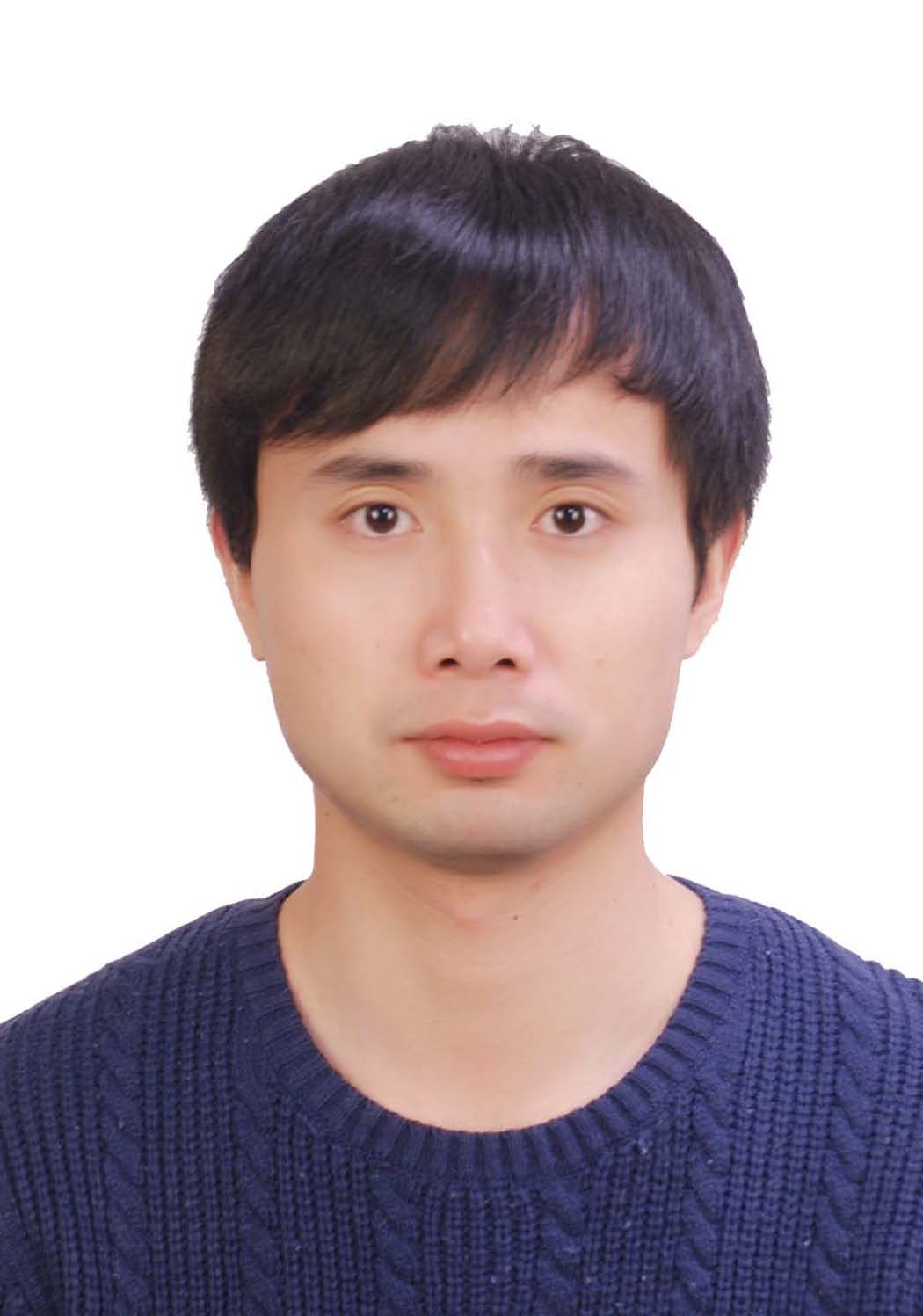}}]{Jiale Cao} received the Ph.D. in information and communication engineering from the Tianjin University, Tianjin, China, in 2018. He is currently an associate professor at the Tianjin University. His research interests include object detection, image/video segmentation, and vision-language learning, in which he has published 30+ papers in top conferences and journals, including IEEE CVPR, IEEE ICCV, ECCV, IEEE TPAMI, IEEE TIP, and IEEE TIFS.
\end{IEEEbiography}

\begin{IEEEbiography}[{\includegraphics[width=1in,height=1.25in,clip,keepaspectratio]{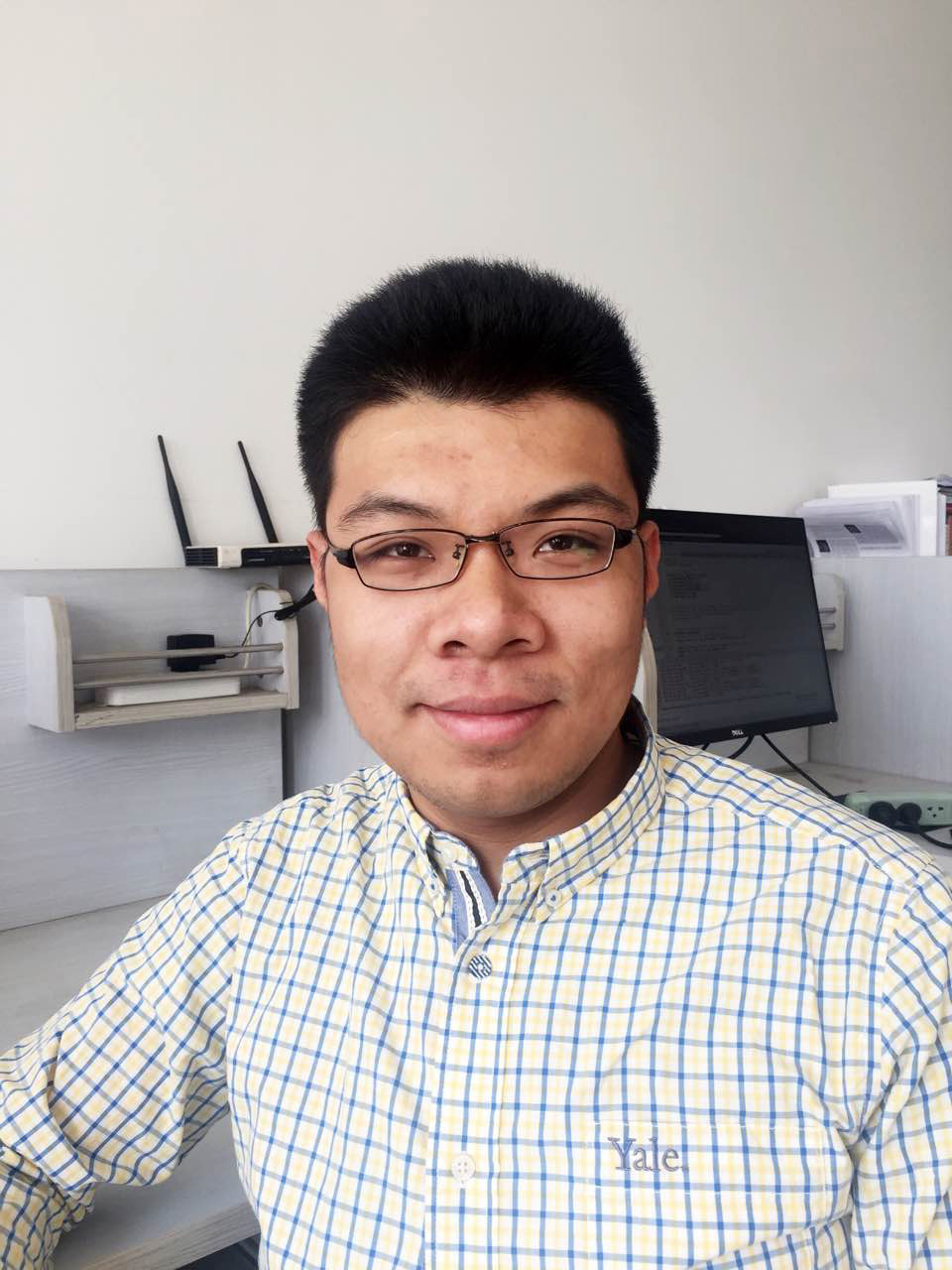}}]{Jin Xie} received the Ph.D. degree in information and communication engineering from Tianjin University, Tianjin, China, in 2021.
	He is currently an associate professor at the Chongqing University.
	His research interests include machine learning and computer vision, in which he has published 20+ papers in IEEE CVPR, IEEE ICCV, ECCV, IEEE TPAMI, and IEEE TIP.
\end{IEEEbiography}

\begin{IEEEbiography}[{\includegraphics[width=1in,height=1.25in,clip,keepaspectratio]{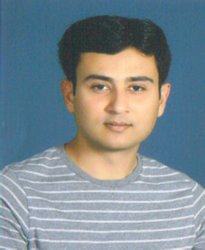}}]{Fahad Shahbaz Khan}  received  the Ph.D. degree in computer vision from the Autonomous University of Barcelona, Barcelona, Spain, in 2011. He is currently a Full Professor and the Deputy Department Chair of Computer Vision with MBZUAI, Abu Dhabi, UAE. He also holds a Faculty Position (Universitetslektor + Docent) with Computer Vision Laboratory, Linköping University, Sweden. From 2018 to 2020, he worked as a Lead Scientist with the Inception Institute of Artificial Intelligence (IIAI), Abu Dhabi. His research interests include a wide range of topics within computer vision and machine learning, such as object recognition, object detection, action recognition, and visual tracking. He has published articles in high-impact computer vision journals and conferences in these areas. He has achieved top ranks on various international challenges (Visual Object Tracking (VOT): 1st 2014 and 2018, 2nd 2015, 1st 2016; VOT-TIR: 1st 2015 and 2016; OpenCV Tracking: 1st 2015; 1st PASCAL VOC 2010).
\end{IEEEbiography}

\begin{IEEEbiography}[{\includegraphics[width=1in,height=1.25in,clip,keepaspectratio]{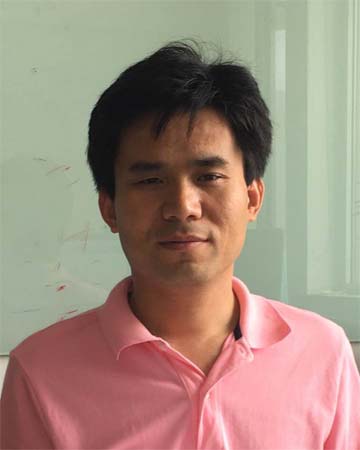}}]{Yanwei Pang} received the Ph.D. degree in electronic engineering from the University of Science and Technology of China. Currently, he is a professor at the Tianjin University and is also the founding director of the Tianjin Key Laboratory of Brain Inspired Intelligence Technology (BIIT). He has published 150+ scientific papers including 50+ IEEE Transactions and 30+ top conferences papers. He was an associate editor of both IEEE T-NNLS and Neural Networks (Elsevier) and a guest editor of Pattern Recognition Letters.
\end{IEEEbiography}








\end{document}